%% file: acl_latex.tex
\documentclass[11pt]{article}
\usepackage[preprint]{acl} 

\usepackage[UTF8, fontset=fandol, scheme=plain]{ctex}

\usepackage{newtxtext}
\usepackage{newtxmath}
\usepackage[protrusion=false,expansion=false]{microtype}
\usepackage{latexsym}
\usepackage{tcolorbox}
\tcbuselibrary{listings, breakable}
\usepackage{listings}

\usepackage[T1]{fontenc}

\usepackage[utf8]{inputenc}

\usepackage{microtype}

\usepackage{inconsolata}

\usepackage{booktabs}
\usepackage{longtable}
\usepackage{multirow}
\usepackage{graphicx}
\usepackage{amsmath}
\usepackage{booktabs}
\usepackage{adjustbox}
\usepackage[table]{xcolor}
\usepackage{booktabs}

\usepackage{lipsum}
\usepackage{xcolor}

\usepackage[colorinlistoftodos]{todonotes}

\title{How Far Are We? Systematic Evaluation of LLMs vs. Human Experts in Mathematical Contest in Modeling}

\author{
  \textbf{Yuhang Liu}\textsuperscript{1,2},
  \textbf{Heyan Huang}\textsuperscript{1,2},
  \textbf{Yizhe Yang}\textsuperscript{1,2},
  \textbf{Hongyan Zhao}\textsuperscript{1},
  \textbf{Zhizhuo Zeng}\textsuperscript{1,2},
  \textbf{Yang Gao}\textsuperscript{1,2,*} \\
  \textsuperscript{1}Beijing Institute of Technology, Beijing, China \\
  \textsuperscript{2}Southeast Academy of Information Technology, Putian, China \\
  \texttt{\{yhliu,hhy63,yizheyang,hongyanzhao,gyang\}@bit.edu.cn}, \texttt{1257781679@qq.com} \\
  \small{Correspondence: \href{mailto:gyang@bit.edu.cn}{gyang@bit.edu.cn}}
}

\begin{document}
\maketitle
\begin{abstract}
Large language models (LLMs) have achieved strong performance on reasoning benchmarks, yet their ability to solve real-world problems requiring end-to-end workflows remains unclear. Mathematical modeling competitions provide a stringent testbed for evaluating such end-to-end problem-solving capability. We propose a problem-oriented, stage-wise evaluation framework that assesses LLM performance across modeling stages using expert-verified criteria. We validate the framework’s reliability by comparing automatic scores with independent human expert judgments on problems from the China Postgraduate Mathematical Contest in Modeling, demonstrating substantially stronger alignment than existing evaluation schemes. Using this framework, we reveal a comprehension–execution gap in state-of-the-art LLMs: while they perform well in early stages such as problem identification and formulation, they exhibit persistent deficiencies in execution-oriented stages including model solving, code implementation, and result analysis. These gaps persist even with increased model scale. We further trace these failures to insufficient specification, missing verification, and lack of validation, with errors propagating across stages without correction. Our findings suggest that bridging this gap requires approaches beyond model scaling, offering insights for applying LLMs to complex real-world problem solving.
\end{abstract}

\input{intro}
\input{method}
\input{related_work}

\input{dataset}

\input{RQ1}

\input{RQ2}

\input{RQ3}

\input{conclusion}
\input{limitation}
\bibliography{custom}

\appendix
\input{appendix/app_main}

\end{document}

%% file: intro.tex
\section{Introduction}
Large language models (LLMs) have demonstrated remarkable performance across a wide range of reasoning tasks ~\cite{deepseekai2025deepseekv32pushingfrontieropen, DBLP:journals/corr/abs-2507-06261, DBLP:journals/corr/abs-2303-08774}, in some cases approaching or surpassing average human performance on standardized benchmarks~\cite{grok, OpenAI2025GPT5SystemCard}. Existing benchmarks~\cite{DBLP:conf/iclr/Yang0HGSHFSL025} are designed to assess individual capabilities, such as mathematical reasoning~\cite{DBLP:conf/nips/HendrycksBKABTS21}, code generation~\cite{DBLP:journals/corr/abs-2506-16395}, or writing~\cite{DBLP:journals/corr/abs-2503-05244}, in isolation. Real-world problem-solving tasks, by contrast, seldom admit such clean decomposition. Instead, they require practitioners to integrate multiple capabilities within a coherent workflow~\cite{DBLP:journals/corr/abs-2509-17677}, including identifying the core problem, making appropriate simplifying assumptions, constructing a suitable model, implementing and solving it computationally, and critically analyzing the results~\cite{xi2025information}. Whether LLMs can perform such integrated, end-to-end problem solving at an expert-level remains an open question.

Mathematical modeling competitions~\cite{banerjee2021mathematical} provide a natural testbed for this question. These competitions present open-ended, real-world problems that demand the full arc of scientific problem solving, from problem framing through implementation to validation, as part of a single problem instance~\cite{bender2000introduction}. Unlike benchmarks with predefined or unique correct answers, modeling problems admit multiple valid approaches, requiring participants to make reasoned choices under uncertainty. Crucially, unlike benchmarks that rely on deterministic automated criteria, solutions in these competitions are assessed by experienced domain experts, making them naturally suited for evaluating expert-level modeling competence.

Although recent studies have shown that LLMs can generate end-to-end solution pipelines and complete mathematical modeling reports~\cite{DBLP:journals/corr/abs-2505-14148, DBLP:journals/corr/abs-2508-09101}, most prior evaluations rely on a fixed set of generic criteria applied uniformly across different modeling problems~\cite{DBLP:journals/corr/abs-2508-09101}. This \emph{problem-agnostic} evaluation paradigm is largely insensitive to the global consistency among modeling objectives, assumptions, decisions, and validation (see Appendix~\ref{app:rubric-example} for concrete examples), and therefore primarily measures report-level presentation quality rather than successful problem solving. As a result, locally plausible yet fundamentally misaligned solutions can still receive high scores. Thus, existing evaluations are incapable of accurately establishing how close such performance is to human expert modeling.

To address this challenge, we propose an evaluation framework that accounts for both problem specificity and the fact that mathematical modeling unfolds through multiple interdependent stages. Specifically, we introduce a \emph{problem-oriental, stage-wise} evaluation framework for full-process mathematical modeling. Our approach replaces static rubrics with dynamically instantiated evaluation schemas that explicitly encode each problem’s objectives, constraints, and success criteria. This enables faithful measurement of modeling competence on real-world tasks and supports interpretable comparisons between LLMs and human expert performance. To ground our evaluation in an expert-level and realistic setting, we use Postgraduate Mathematical Contest in Modeling (PMCM) as our primary evaluation benchmark because it represents genuine expert-level difficulty, with only about 1.0 to 1.5\% of participants achieving gold medals~\cite{GMCM2025Awards}. We investigate the following research questions:

\paragraph{RQ1: Does the framework align with expert judgment?}
By implementing expert-verified, subtask- and stage-specific evaluation criteria, our proposed framework generates substantially more reliable and discriminative scores compared to problem-agnostic rubrics. Correlation analysis against expert judgments reveals consistently stronger alignment across all evaluated modeling stages, thereby confirming the enhanced evaluation validity of our approach.

\paragraph{RQ2: Where do LLMs fall short across modeling stages?}
Building upon the proposed evaluation framework, we conduct a stage-wise analysis of LLM-generated mathematical modeling reports under a unified 10-point scoring scheme, encompassing the modeling workflow from problem framing to result validation. Our findings reveal stage-dependent performance disparities. LLMs demonstrate near-expert performance in problem understanding and initial framing. Conversely, performance in the core modeling stage, encompassing assumption formulation, model construction, and model solving, remains at an intermediate level of approximately 4–6 out of 10. More critically, LLMs exhibit poor performance in code implementation and result validation, with scores typically falling within the 2–3 range. These performance patterns persist even for state-of-the-art large-scale models, suggesting that increased model scale alone may not resolve the challenges associated with full-process mathematical modeling.

\paragraph{RQ3: What causes LLM failures in modeling?}
Our stage-wise failure analysis reveals that performance degradation is primarily attributed to execution-level deficiencies rather than erroneous problem understanding or modeling intent. In the development of assumptions and model construction, failures predominantly arise from inadequate specification, justification, and verification, even when high-level modeling choices appear reasonable. These early-stage shortcomings are rarely revisited in subsequent stages, resulting in downstream failures characterized by missing verification, incomplete procedures, and a lack of validation. Concurrently, these findings suggest that LLMs encounter less difficulty in generating plausible modeling concepts than in executing, validating, and refining them across stages.

In summary, we propose a reliable, problem-oriented evaluation framework that grounds assessment criteria in problem semantics and modeling stages, enabling faithful measurement of mathematical modeling competence. Using this framework, we conduct a systematic comparison between LLMs and human experts, revealing substantial performance gaps particularly in execution-oriented stages such as model solving, code implementation, and result analysis. We further trace these gaps to specific failure patterns including insufficient specification, missing verification, and lack of validation, which persist across stages without corrective feedback. These findings suggest that future improvements should focus on enhancing procedural rigor and stage-wise consistency rather than solely scaling model size, pointing toward process-aware approaches that prioritize executability and validation throughout the modeling pipeline.


%% file: method.tex
\section{Problem-Oriented Stage-Wise Evaluation Framework}
\label{sec:eval_instantiation}

\begin{figure*}[tbp]
    \centering
    \includegraphics[height=4cm,width=0.85\textwidth]{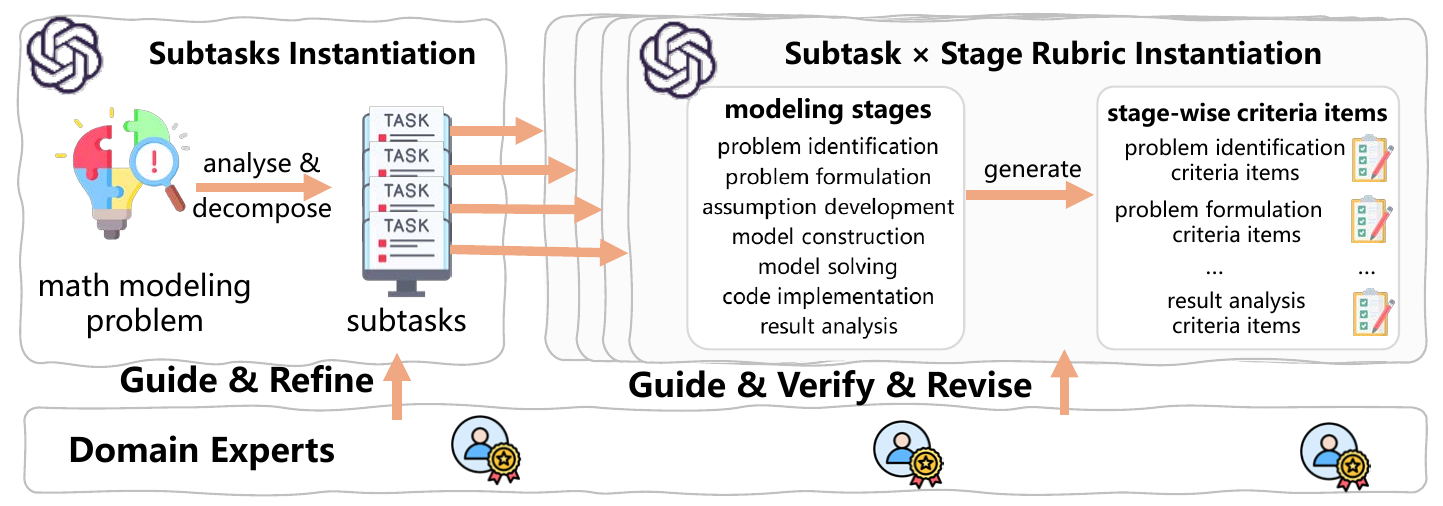}
    \caption{Problem-oriented, stage-wise evaluation framework. An LLM first decomposes a mathematical modeling problem into a set of subtasks, which are guided and refined by domain experts to preserve the original intent. For each verified subtask, a stage-aware evaluation rubric is instantiated by conditioning on the major stages of the modeling process. For each subtask--stage pair, the LLM generates concrete evaluation criteria under expert guidance, which serve as the atomic units for scoring and are further verified and revised. The resulting subtask--stage rubric defines a fixed, problem-specific evaluation structure applied uniformly across models.}
    \label{fig:rq1-flow}
\end{figure*}
\label{sec:results-rq1}

To accurately assess the performance of mathematical modeling, we introduce a \textit{problem-oriented, stage-wise} evaluation framework. The framework operates in two phases. First, we decompose each mathematical modeling problem into a set of subtasks, where each subtask corresponds to a self-contained modeling requirement. Second, for each subtask, we construct evaluation criteria aligned with canonical modeling stages that span the full modeling workflow from problem understanding to result validation. Both phases incorporate expert-in-the-loop verification to ensure reliability, where domain experts review and refine the outputs before they are applied uniformly across all evaluated models.

\subsection{Subtask Decomposition}

A single mathematical modeling problem typically comprises multiple modeling requirements that cannot be reliably evaluated as a single unit. To avoid conflating coverage of requirements with modeling quality, we decompose each problem into a set of distinct subtasks. Each subtask corresponds to a self-contained modeling requirement that must be addressed through a complete modeling pipeline, including formulation, assumption specification, model construction, solution, implementation, and analysis. We first prompt an LLM to analyze the problem and generate an initial set of subtasks following the structured decomposition strategy by~\citet{DBLP:journals/corr/abs-2508-09101}. Domain experts then examine whether the resulting subtasks collectively capture the key information, objectives, and requirements expressed in the original problem description. Based on expert feedback, the decomposition is iteratively refined until the subtasks provide a faithful and stable representation of the modeling task. The verified subtasks serve as a shared basis for both mathematical modeling report generation and evaluation, ensuring that all evaluated models are assessed against the same problem structure. Concrete examples of subtask decomposition are provided in Appendix~\ref{app:subtask-examples}.

\subsection{Stage-wise Evaluation Criteria Construction}

For each verified subtask, we construct evaluation criteria aligned with canonical modeling stages. The modeling process is organized into seven stages that span the full workflow, including {\em problem identification, problem formulation, assumption development, model construction, model solving, code implementation}, and {\em result analysis}, following standard decompositions of mathematical and scientific modeling practice~\citep{bender2000introduction,banerjee2021mathematical}. Detailed definitions of each stage are provided in Appendix~\ref{app:stage-definitions}. By explicitly conditioning evaluation on both the subtask and the modeling stage, we enable fine-grained and diagnostically meaningful assessment.

For each subtask--stage pair, we prompt an LLM to generate concrete evaluation criteria grounded in the subtask's objectives, inputs, constraints, and expected outputs. Each criterion specifies a necessary condition for successful completion of the corresponding stage, formulated in a task-specific and checkable manner. All criteria are generated with expert-in-the-loop support: when necessary, domain experts intervene to refine the prompting instructions used by the LLM, guiding the generation toward task-faithful and stage-appropriate criteria. The resulting criteria are then verified to ensure correctness, completeness, and alignment with expert modeling practice. A concrete example of stage-wise criteria for a real modeling subtask is provided in Appendix~\ref{app:subtask-examples}.

Each verified criterion serves as an atomic unit for scoring. Model-generated reports are evaluated separately against every criterion associated with a given subtask and stage, yielding a structured score profile across the full modeling pipeline. Because the same verified criteria are applied uniformly to all models, this approach ensures consistent and comparable evaluation.

%% file: related_work.tex
\section{Related Work}

\paragraph{Evaluation of LLMs on Reasoning Tasks}
A large body of work evaluates LLMs on reasoning-intensive benchmarks, including mathematical problem solving, logical reasoning, scientific question answering, and code generation~\cite{AIME,DBLP:journals/corr/abs-2505-14615,song2025evaluatinglargelanguagemodels,DBLP:conf/iclr/JimenezYWYPPN24}. These evaluations typically focus on answer correctness or the coherence of intermediate reasoning steps under well-defined problem settings. While recent models achieve strong performance on narrowly scoped benchmarks, such evaluations largely isolate individual reasoning components and do not assess a model’s ability to manage the full problem-solving lifecycle.

\paragraph{LLMs for End-to-End and Open-Ended Problem Solving}
Recent work has applied LLMs to more open-ended, multi-stage tasks such as scientific analysis, software development, and full mathematical modeling workflows~\cite{DBLP:journals/corr/abs-2505-09970,DBLP:journals/corr/abs-2511-08151,DBLP:journals/corr/abs-2506-14683,qian2025modelingagentbridgingllmsmathematical}. In these settings, evaluation remains largely coarse-grained, relying on holistic rubrics or preference judgments~\cite{DBLP:conf/kdd/MohammadiLLY25,DBLP:conf/acl/YanFYX00Z25,wang2025ariseagenticrubricguidediterative}. As a result, existing evaluations provide limited insight into stage-wise execution and coordination, motivating the need for more fine-grained, problem-conditioned assessment frameworks.

%% file: dataset.tex
\section{Experimental Setup}
\label{sec:dataset-setup}

\subsection{Dataset and Preprocessing}
Our benchmark consists of graduate-level mathematical modeling problems drawn from recent editions of the China PMCM. Each problem is originally released as a multi-page PDF document and is designed for postgraduate participants, requiring expert-level modeling competence that integrates problem formulation, assumption design, model construction, computational implementation, and result validation within a single open-ended task, while involving dense symbolic expressions, structured mathematical arguments, and rich real-world context. As such, they represent a stringent testbed for expert-level, end-to-end mathematical modeling.

The final dataset comprises 97 problems, each spanning multiple pages of text, formulas, figures, and tables. To obtain a canonical machine-readable representation, we first process the original PDF files using a carefully designed prompting pipeline with \texttt{Qwen2.5-VL-32B}~\cite{DBLP:journals/corr/abs-2502-13923}, converting page-level PDF content into LaTeX. The resulting outputs are then stitched, normalized, and merged into a single LaTeX file per problem. All converted content is subsequently reviewed and corrected by the authors to ensure mathematical correctness, structural fidelity, and consistency with the original problem statements. We additionally extract basic metadata, including topical tags and structural attributes, to facilitate downstream analysis. Full preprocessing details are provided in the Appendix~\ref{app:data_preprocess}.

\subsection{Generation Protocol and Models}
To ensure a consistent and comparable generation process across LLMs, we adopt a unified mathematical modeling agent framework for all experiments, fixing the subtask structure for each problem so that the same set of subtasks, ordering, input formats, and output constraints are used for all models. Specifically, we follow the report-generation framework in~\citet{DBLP:journals/corr/abs-2505-14148}, which generates a complete modeling report for each problem by first decomposing the problem into a sequence of subtasks, then prompting an LLM to solve each subtask in turn, and finally aggregating the subtask outputs into a coherent final report. For our setting, the original English prompts are translated into Chinese to match the China PMCM problems. The subtask specifications are derived from the problem-conditioned schema described in Section~\ref{sec:eval_instantiation}, ensuring that the generation pipeline is aligned with the evaluation decomposition while keeping the LLMs as the sole varying component.

Within this fixed agent protocol, we evaluate a diverse set of contemporary instruction-tuned LLMs to examine performance across different parameter scales and reasoning capabilities. Our evaluation includes multiple variants from the \texttt{Qwen} family ranging from 7B to 235B parameters~\cite{DBLP:journals/corr/abs-2412-15115, DBLP:journals/corr/abs-2505-09388}, the \texttt{DeepSeek-V3.2-Instruct} and \texttt{DeepSeek-V3.2-Thinking} models from the DeepSeek-V3.2-Exp series~\cite{deepseekai2024deepseekv32}, and \texttt{o4-mini}~\cite{o4mini}.

%% file: RQ1.tex

\section{Evaluation Reliability: Alignment with Expert Judgment}

\begin{table}[t]
\centering
\begin{tabular}{l cc}
\toprule
 & Baseline & Ours \\
\midrule
ICC(2,1) w.r.t.\ Expert & 0.012 & \textbf{0.673} \\
\bottomrule
\end{tabular}
\caption{Agreement between independent expert overall judgments and automatic scores.}
\label{tab:expert-alignment}
\end{table}

Before using our framework to evaluate LLMs, we first validate its reliability. A reliable evaluation framework should produce scores that align with human expert judgment and capture genuine problem-solving quality, including satisfaction of modeling objectives, adherence to problem constraints, and coherence across modeling stages, rather than superficial aspects of report presentation. 

To assess reliability, we recruit medal-winning mathematical modeling students as domain experts to provide independent judgments. These experts evaluate each LLM-generated report at the level of subtask- stage pairs, scoring against the same criterion items used in our automatic evaluation. Their judgments serve as the reference signal for measuring evaluation quality. We then compare the agreement between these expert scores and automatic scores produced by two evaluation schemes. The first is our proposed framework, which evaluates reports at the level of subtask and stage criterion items. The second is a baseline following the rubric of~\citet{DBLP:journals/corr/abs-2505-14148}, which evaluates each modeling report along coarse-grained dimensions such as problem analysis, modeling rigor, practicality and scientificity, and result and bias analysis (see Table~\ref{tab:case-overview}). We quantify agreement using ICC(2,1)\footnote{The detailed definitions and implementation are provided in Appendix~\ref{app:icc}}, which measures absolute agreement under a two-way random-effects model.

\begin{figure}[t]
    \centering
    \includegraphics[width=0.45\textwidth]{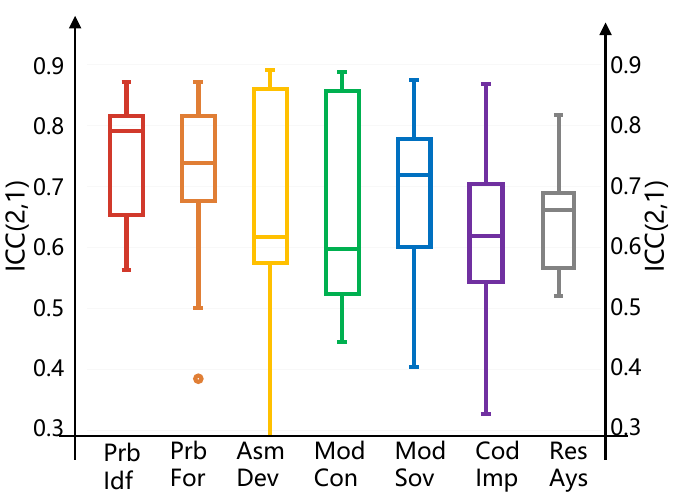}
    \caption{\textbf{Stage-wise agreement} between automatic and human expert scores measured by ICC(2,1). Higher values indicate stronger alignment with expert assessment. The boxplots correspond to individual modeling stages (Prb Idf, Prb Frm, Asm Dev, Mod Con, Mod Sol, Cod Imp, Res Ays), following the standard mathematical modeling workflow.}
    \label{fig:rq1-stage}
\end{figure}
\paragraph{Stage-wise framework matches expert judgment.}
We evaluate the reliability of the proposed framework by measuring its agreement with expert judgments at two complementary levels. At the {\em report level}, we compare both evaluation schemes against independent expert judgments. For our framework, we aggregate criterion-level scores across stages within each subtask and then across subtasks to obtain a single report-level score. For the baseline, we directly average scores across its predefined coarse-grained dimensions. Table~\ref{tab:expert-alignment} reports agreement measured by ICC(2,1). The baseline attains a value of only 0.012, indicating virtually no alignment with expert judgments. In contrast, our framework achieves an ICC of \textbf{0.673}, demonstrating substantially stronger consistency with expert assessments. This result confirms that fine-grained, problem-oriented evaluation captures modeling quality far more faithfully than coarse-grained report-level rubrics. We further examine reliability at the {\em criterion level}, we further examine whether this reliability holds across different modeling stages. Figure~\ref{fig:rq1-stage} reports ICC(2,1) values computed from criterion-level scores within each stage. Across all stages, we observe consistently high agreement between automatic and expert scores, with median ICC values exceeding commonly accepted thresholds for good reliability. Notably, strong agreement extends beyond early, language-driven stages such as problem identification and formulation to technically demanding stages including model construction, model solving, and code implementation. These results indicate that our framework yields stable and reproducible scores throughout the full modeling pipeline.

\begin{figure}[t]
    \centering
    \includegraphics[width=0.9\linewidth]{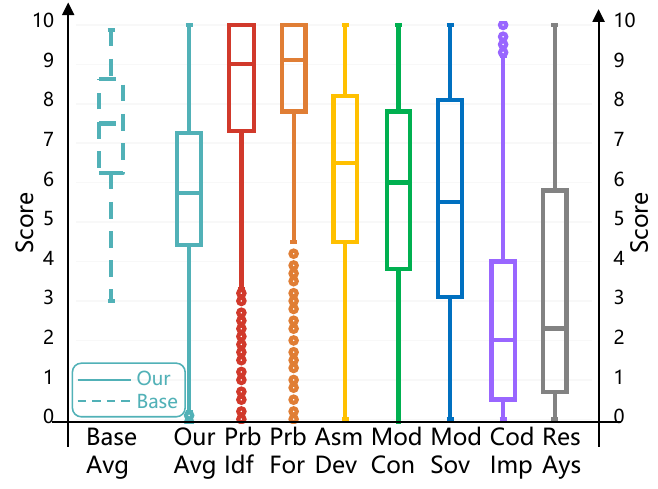}
    \caption{Distribution of \textbf{evaluation scores} under the baseline rubric and our framework. \textbf{Base Avg} denotes the report-level baseline score, and \textbf{Our Avg} denotes the overall score under our problem-oriented, stage-aware framework. The remaining boxplots correspond to individual modeling stages (Prb Idf, Prb Frm, Asm Dev, Mod Con, Mod Sol, Cod Imp, Res Ays).}
    \label{fig:rq1-stage-boxplot}
\end{figure}

\paragraph{Stage-wise framework can distinguishes solution quality.}
We next examine whether our framework provides more discriminative assessment than the baseline. A valid evaluation should distinguish solutions that differ in their fulfillment of problem-specific modeling requirements, rather than collapsing them into a narrow score range. Figure~\ref{fig:rq1-stage-boxplot} compares the score distributions under the baseline rubric and our framework. Baseline scores are heavily concentrated in the mid-range of approximately 6.5 to 8.5, indicating that reports frequently receive similar moderate to high scores regardless of actual modeling quality. This concentration persists even for reports that exhibit clear deficiencies in specific modeling stages. We conduct a diagnostic case analysis in Appendix~\ref{app:case}, demonstrating that high baseline scores can coexist with violations of problem-grounded necessary modeling conditions. In contrast, scores produced by our framework exhibit substantially greater dispersion across stages and subtasks. Because evaluation is performed at the subtask-stage level, deficiencies in specific modeling components cannot be offset by strengths in unrelated parts of the report. Failures in technically demanding stages such as model construction, model solving, or code implementation are explicitly reflected in lower scores, even when earlier stages are well executed. This structure prevents systematic score inflation and enables discriminative assessment of whether the modeling task has been substantively completed.

%% file: RQ2.tex
\section{Stage-wise Performance: The Comprehension-Execution Divide}

Having established the reliability of our evaluation framework, we now apply it to assess how far current LLMs remain from human expert performance in mathematical modeling. Rather than treating modeling ability as a single aggregate outcome, we conduct a stage-wise analysis to identify where LLMs systematically fall short. We further examine whether these performance gaps can be mitigated through model scaling.


\begin{figure}[tbp]
    \centering
    \includegraphics[width=0.45\textwidth,height=5.8cm]{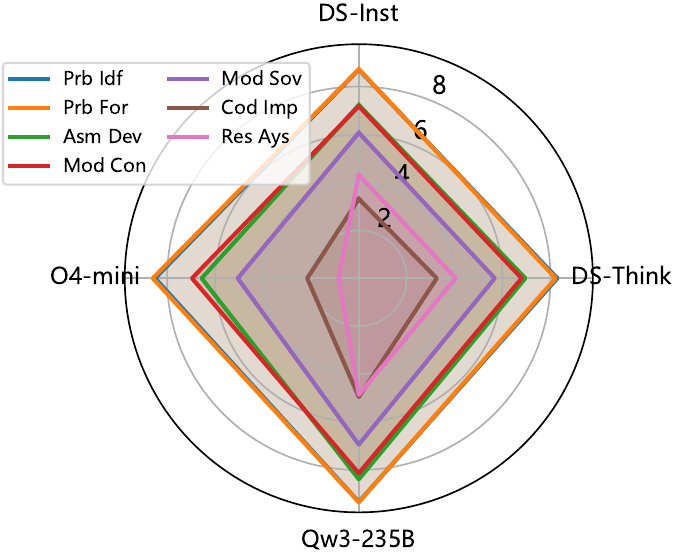}
    \caption{Stage-wise performance of four LLMs across the mathematical modeling pipeline. DS-Inst and DS-Think denote two DeepSeek-V3.2 variants, and Qw3-235B denotes a Qwen3-235B-Instruct model. Scores are averaged over reports and subtasks for each stage.}
    \label{fig:rq1-radar-overall}
\end{figure}

\subsection{The Comprehension-Execution Gap}

\paragraph{LLMs excel at comprehension but struggle with execution.}
We first examine how LLM performance varies across different modeling stages. For each modeling report, we compute a stage-level score by averaging scores across all subtasks associated with that stage. We then aggregate these report-level scores across all evaluated reports to obtain model-level stage scores.  Figure~\ref{fig:rq1-radar-overall} reports mean stage-level scores under our framework. 
Across all models, performance exhibits a clear monotonic decline as the modeling process progresses. Models achieve their highest scores in early, comprehension-oriented stages such as problem identification and problem formulation, but scores decrease steadily in subsequent stages. By the time models reach execution-heavy stages, most notably model solving, code implementation, and result analysis, average scores often fall to around 5.0 or below. This pattern indicates that, even when earlier reasoning steps are handled adequately, current LLMs struggle to sustain correctness and completeness through the later stages required for end-to-end mathematical modeling.

\paragraph{Understanding problems differs from modeling them.}
High performance in early stages primarily reflects the ability to interpret problem statements, restate objectives, and organize inputs into a coherent structure. However, these capabilities do not translate into reliable performance in stages that require producing checkable solutions, executing computational procedures, and validating results. The sharp score drop observed at the model-solving stage marks a critical transition point, beyond which purely language-driven reasoning becomes insufficient to support full-process modeling. As a result, current LLMs, when used in isolation, remain unable to consistently complete mathematical modeling tasks end to end.

\begin{figure}[tbp]
    \centering
    \includegraphics[width=0.45\textwidth,height=5.5cm]{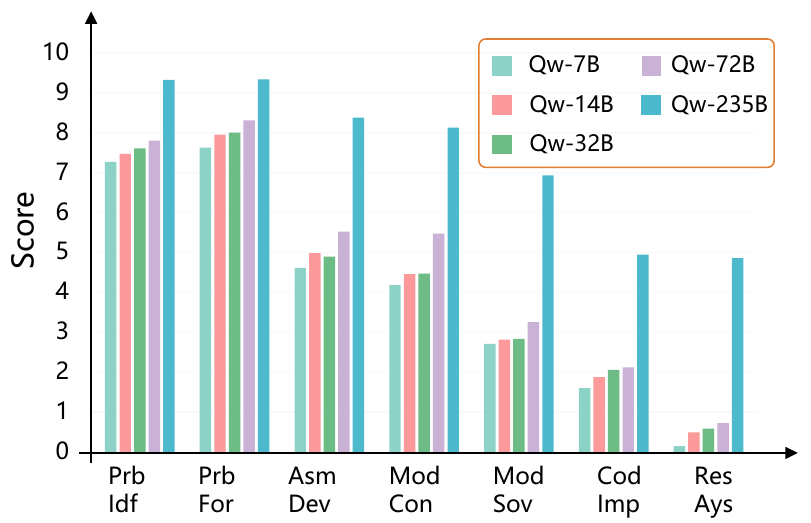}
    \caption{Stage-wise performance under within-family scaling for Qwen models. Bars correspond to \texttt{Qw-7B}--\texttt{Qw-72B} (Qwen2.5) and \texttt{Qw-235B} (Qwen3-235B).}
    \label{fig:rq1-qwen-scaling}
\end{figure}

\subsection{Scaling Alone Is Not Enough}
Given the substantial comprehension--execution gap identified above, a natural question arises: can these performance gaps be closed by scaling model size? To investigate this, we analyze models within the Qwen family, ranging from 7B to 235B parameters. Figure~\ref{fig:rq1-qwen-scaling} presents the stage-wise performance across different model scales.

\paragraph{Scaling improves comprehension but with diminishing gains.}
Figure~\ref{fig:rq1-qwen-scaling} shows that performance in early modeling stages such as problem identification and problem formulation increases steadily as model scale grows from 7B to 72B parameters. This indicates that larger models are more reliable at understanding problem statements, extracting objectives, and organizing the problem structure. However, the magnitude of improvement tapers off as scale increases. While the jump from smaller to mid-sized models yields noticeable gains, further scaling leads to progressively smaller improvements. These stages primarily depend on linguistic comprehension and structural organization, which benefit from scale but do not require fundamentally new capabilities. As a result, early-stage modeling performance improves with scale yet saturates relatively quickly.

\paragraph{Scaling fails to close the execution gap.}
In contrast, execution-oriented stages including model solving, code implementation, and result analysis exhibit little consistent improvement as model size increases from 7B to 72B. Scores in these stages remain low and relatively flat across scales, indicating that additional parameters alone do not translate into better execution, verification, or validation in practice. These stages require tightly coupled chains of reasoning, precise procedural execution, and empirical grounding, none of which appear to develop reliably through scaling alone. This persistent gap suggests that full mathematical modeling competence is not a linear extension of language understanding, and that improving execution-oriented performance may require approaches beyond simply increasing model size.

%% file: RQ3.tex
\begin{table*}[t]
\centering
\small

\label{tab:failure-causes}
\begin{tabular}{p{3.5cm} p{12cm}}
\toprule
Stage & Common Failure Causes \\
\midrule

Assumption Development &
Assumptions Not Checked; Missing Assumption Conditions; Assumption Impact Not Discussed; 
Hidden Idealized Assumptions; Unrealistic Assumptions  \\

Model Construction &
Incomplete Model Structure; Missing Model Derivation; Unclear Variables or Symbols;
Solvability Not Justified; Model Deviates from Task Goal; Model-Assumption Conflict\\

Model Solving &
No Checkable Solution; Key Solution Steps Missing; 
Solution Not Verified; Numerical Stability Not Analyzed; Computationally Infeasible; Inappropriate Solution Method\\

Code Implementation &
Cannot be executed; Results Not Reproducible; No Code;
Engineering or Numerical Risks; Code--Model Mismatch  \\

Result Analysis &
No Meaningful Results; No Validation or Comparison; Sensitivity Not Analyzed;
Conclusions Not Supported; Results Miss the Goal; Limits Not Discussed \\

\bottomrule
\end{tabular}
\caption{Stage-specific failure causes across the mathematical modeling workflow. For each stage, failure causes are ordered from left to right by decreasing prevalence, based on their frequency of occurrence in reviewer comments associated with low-scoring cases. The detailed  aggregation procedure of these frequencies is described in Appendix~\ref{app:rq3_supple}.}
\label{tab:failure-causes}
\end{table*}

\section{Failure Analysis: Execution Deficiencies Dominate}
\label{sec:stage-wise-failure-analysis}

Having identified substantial performance gaps in execution-oriented stages, we now investigate why these failures occur. We perform a stage-wise failure analysis by examining all subtasks whose stage-level score falls below 8 points. For these low-scoring cases, we collect the associated evaluation comments and prompt an LLM to group recurring or semantically equivalent comments into coherent categories. The LLM then synthesizes each category into a representative failure cause for the corresponding stage. Table~\ref{tab:failure-causes} summarizes the resulting failure causes, with detailed operational definitions provided in Appendix~\ref{app:failure-causes}. Within each stage, causes are ordered by their relative prevalence in reviewer comments, highlighting the dominant failure patterns at different points in the modeling pipeline.


\paragraph{Reasonable ideas, poor execution.}
Across all stages, the most prevalent failure causes are execution-level deficiencies rather than fundamentally flawed modeling ideas. LLMs generally propose reasonable high-level approaches, but fail to carry them out with sufficient rigor. In assumption development, models frequently state plausible assumptions without checking their validity, specifying necessary conditions, or analyzing their impact on downstream results. In model construction, models often produce structures that appear aligned with task objectives but lack complete derivations, clear variable definitions, or justification for solvability. These patterns reveal a consistent gap between modeling intent and modeling execution: LLMs can identify what to do, but struggle with how to do it properly.


\paragraph{Errors propagate without correction.}
Beyond individual stage failures, we observe that deficiencies introduced in earlier stages are rarely corrected in subsequent ones. Later stages such as model solving, code implementation, and result analysis are dominated by failures including solution not verified, no checkable solution, results not reproducible, and no validation or comparison. These downstream failures are consistent with earlier-stage issues such as underspecified assumptions and incomplete model formulations. This pattern suggests that LLMs treat intermediate artifacts as fixed inputs rather than revisiting or revising them based on later findings. As a result, early-stage weaknesses propagate and compound throughout the modeling pipeline.

Overall, these failure patterns reveal that current LLMs struggle to maintain procedural rigor across the full mathematical modeling workflow. While LLMs often exhibit reasonable modeling intent in early stages, they fail to sustain this quality through execution and validation. The core challenge is not generating plausible ideas, but carrying them out with sufficient specification, verification, and validation at each stage. These findings suggest that future improvements should target process-level consistency and self-correction mechanisms rather than focusing solely on enhancing individual stage capabilities.


%% file: conclusion.tex
\section{Conclusion}

This work investigates the extent to which LLMs can perform full-process mathematical modeling under expert-level competition settings. We propose a problem-oriented, stage-wise evaluation framework that grounds assessment criteria in problem semantics and modeling stages, enabling more reliable evaluation than problem-agnostic rubrics. Using this framework, we reveal a substantial comprehension–execution gap between LLMs and human experts, where LLMs perform well in understanding problems but struggle significantly in solving models, implementing code, and validating results. We further trace these gaps to specific failure patterns, showing that LLMs can generate reasonable modeling ideas but fail to carry them out with sufficient specification, verification, and validation, and that early-stage errors propagate throughout the pipeline without correction. These findings fundamentally reshape our understanding of how to advance LLMs toward expert-level problem solving. Rather than relying on model scaling alone, future progress requires new paradigms that bridge the comprehension-execution gap, such as process-aware reasoning and stage-wise self-correction. 


%% file: limitation.tex
\section{Limitations}

Our study is subject to the limitation that the proposed evaluation framework is instantiated and validated on mathematical modeling problems from the Postgraduate Mathematical Contest in Modeling (PMCM). While PMCM provides a realistic and expert-level setting that requires full-process reasoning, it represents a specific class of open-ended tasks with well-established modeling conventions and evaluation practices. As a result, the stage decomposition and evaluation criteria adopted in this work may not directly transfer to other end-to-end problem-solving domains with different structures or validation norms. In addition, the instantiation of problem-conditioned, stage-wise evaluation criteria relies on expert guidance to ensure faithful coverage of problem requirements and modeling semantics, which may limit scalability in settings where such expertise is unavailable. Extending the framework to a broader range of tasks while reducing reliance on expert intervention remains an important direction for future research.

%% file: appendix/app_main.tex
\onecolumn
\appendix
\label{sec:appendix}
\section{Ethical Considerations}
We recruited three experts to perform the annotation tasks. All annotators are students majoring in mathematics and have received medals in mathematical modeling competitions, indicating recognized expertise in mathematical modeling. Among them, two are PhD candidates and one is a master’s student. All annotation procedures were conducted in accordance with institutional ethical guidelines and were approved by the Institutional Review Board (IRB).

\input{appendix/subtask}

\input{appendix/stage_definition}

\input{appendix/data_preprocess}

\input{appendix/ICC}

\input{appendix/case}

\input{appendix/stage_faliure_causes}

\input{appendix/RQ3_supplementary}

\input{appendix/problem_analysis}

\input{appendix/prompts/criteria}
\input{appendix/prompts/eval_prompt}

%% file: appendix/subtask.tex
\section{Examples of Subtask Instantiation from a Modeling Problem}
\label{app:subtask-examples}

To illustrate what constitutes a \emph{subtask} in our evaluation framework, we present four representative subtasks instantiated from a real mathematical modeling problem on maneuvering target tracking. The purpose of these examples is not to reproduce a full solution, but to demonstrate how complex problem requirements are decomposed into self-contained modeling subtasks that admit stage-wise evaluation.

\subsection{Original Problem Context (Excerpt).}
The problem concerns tracking maneuvering targets using radar measurements. The provided datasets contain asynchronous observations from one or multiple radar sensors, with each observation specified in polar coordinates (range, azimuth, elevation, timestamp, and sensor identifier). The overall task involves coordinate transformation, uncertainty modeling, dynamic state estimation under maneuver, data association in multi-target scenarios, and long-horizon trajectory prediction.

\subsection{Subtask 1: Measurement Preprocessing and Observation Uncertainty Modeling.}

\emph{Subtask Definition.}
Transform raw radar observations from polar coordinates into a unified Cartesian coordinate system, quantify and propagate measurement uncertainty, handle asynchronous timestamps across sensors, detect and mark outliers, and output a time-ordered observation sequence suitable for downstream filtering and data association.

\emph{Subtask Scope and Inputs.}
The inputs to this subtask include raw radar measurements $(r, \mathrm{az}, \mathrm{el}, t, s)$, sensor metadata (sensor positions in geodetic or Earth-centered coordinates), and sensor-specific measurement error statistics (range and angular uncertainties). The subtask assumes known sensor locations and valid timestamps, while allowing observations to be asynchronous and heterogeneous across sensors.

\emph{Expected Outputs.}
The output is a structured, time-ordered observation sequence. Each record contains:
(i) a Cartesian position estimate in a unified reference frame,
(ii) an associated measurement covariance matrix representing propagated uncertainty,
(iii) the observation timestamp and sensor identifier,
and (iv) auxiliary metadata such as sensor geometry and quality or anomaly flags.

\emph{Role in the Modeling Pipeline.}
This subtask serves as the interface between raw sensor data and all subsequent modeling stages. Errors introduced here (e.g., incorrect coordinate conventions or missing uncertainty propagation) systematically affect downstream estimation and cannot be corrected later, making it a necessary and independently evaluable modeling requirement.

\subsection{Subtask 2: Dynamic Model Design and State Estimation under Maneuver.}

\emph{Subtask Definition.}
Design appropriate target motion models and estimate the time-varying target state (e.g., position and velocity) from noisy, nonlinear observations, while adapting to potential maneuvering behavior.

\emph{Subtask Scope and Inputs.}
The inputs include the preprocessed observation sequence from Subtask~1, together with modeling assumptions about target dynamics (e.g., constant velocity, constant acceleration, or stochastic acceleration models). Measurement functions and noise models are assumed known from preprocessing.

\emph{Expected Outputs.}
The output consists of a time-indexed sequence of state estimates and associated uncertainty measures (state covariance matrices). When multiple motion hypotheses are considered, the output may additionally include model probabilities or diagnostic indicators reflecting detected maneuvers.

\emph{Role in the Modeling Pipeline.}
This subtask constitutes the core estimation component of the tracking system. It translates observational data into latent target states and determines how quickly and reliably the tracker responds to maneuvers. Its performance directly affects the quality of subsequent data association and prediction tasks.

\subsection{Subtask 3: Data Association and Track Management in Multi-Target Scenarios.}

\emph{Subtask Definition.}
Associate incoming observations with existing target tracks, manage track initiation and termination, and maintain track continuity under ambiguous, missing, or overlapping measurements in multi-target environments.

\emph{Subtask Scope and Inputs.}
The inputs include predicted target states and uncertainties from Subtask~2, together with the current set of observations and their covariances from Subtask~1. The subtask may further rely on assumed detection probabilities and clutter characteristics.

\emph{Expected Outputs.}
The output includes updated target tracks with revised state estimates, measurement-to-track association decisions or probabilities, and track-level status indicators (e.g., confirmed, tentative, coasting, or terminated). In probabilistic formulations, association likelihoods or marginal probabilities are also produced.

\emph{Role in the Modeling Pipeline.}
This subtask ensures identity consistency over time and prevents error accumulation caused by incorrect measurement-to-track assignments. Failures at this stage can lead to track swaps, fragmentation, or loss, which cannot be remedied by later prediction or smoothing.

\subsection{Subtask 4: Trajectory Prediction and Outcome Estimation.}

\emph{Subtask Definition.}
Predict future target motion beyond the observation window based on current state estimates and dynamic models, and estimate task-specific outcomes such as future trajectory envelopes or landing locations, together with associated uncertainty.

\emph{Subtask Scope and Inputs.}
The inputs consist of the latest state estimates and uncertainties from Subtask~2 (and, where relevant, track hypotheses from Subtask~3), along with assumed future dynamics and environmental constraints. This subtask may incorporate stochastic simulation or uncertainty propagation methods.

\emph{Expected Outputs.}
The output includes predicted trajectories or outcome distributions (e.g., expected landing points with confidence regions), temporal forecasts, and uncertainty summaries. In addition, computational cost or complexity estimates may be reported to assess feasibility under real-time constraints.

\emph{Role in the Modeling Pipeline.}
This subtask translates state estimation results into actionable predictions and decision-relevant quantities. It is the final consumer of all upstream modeling outputs and provides the basis for interpretation, planning, or strategic analysis.

\paragraph{Why These Constitute Subtasks.}
Each of the above subtasks addresses a distinct modeling requirement implied by the original problem, admits a complete modeling pipeline from formulation to output, produces well-defined artifacts consumed by downstream stages, and can be meaningfully evaluated using stage-aware criteria. Together, they illustrate how a complex modeling problem can be systematically decomposed into independently evaluable subtasks within our framework.

%% file: appendix/stage_definition.tex
\section{Stage Definitions and Example}
\label{app:stage}

\subsection{Stage Definitions}
\label{app:stage-definitions}
\paragraph{Problem Identification.}
Problem identification refers to the initial understanding of the real-world task, including recognizing the core objective, relevant entities, and success criteria implied by the problem description. At this stage, the modeler determines what is to be predicted, optimized, or explained, and distinguishes essential requirements from contextual background. Accurate problem identification is a prerequisite for all subsequent modeling decisions; misunderstandings at this stage fundamentally alter the nature of the task and cannot be compensated for by later technical sophistication.

\paragraph{Problem Formulation.}
Problem formulation translates the identified real-world task into a formal modeling problem. This includes defining inputs and outputs, specifying target variables, clarifying constraints, and determining the overall modeling scope. While problem identification focuses on understanding \emph{what} is required, formulation focuses on deciding \emph{how} the problem should be represented in analytical or mathematical terms. Errors at this stage often lead to internally consistent but misaligned solutions that fail to address the original task.

\paragraph{Assumption Development.}
Assumption development involves explicitly stating and justifying the simplifying assumptions necessary to make the problem tractable. These assumptions may concern data quality, noise characteristics, independence, stationarity, linearity, or domain-specific conditions. Assumptions play a central role in mathematical modeling by connecting real-world complexity to formal representations. Evaluating this stage is essential, as unstated, unjustified, or violated assumptions can invalidate an otherwise well-constructed model.

\paragraph{Model Construction.}
Model construction concerns the selection and specification of mathematical structures that operationalize the formulated problem under the stated assumptions. This includes defining equations, objective functions, probabilistic models, or algorithmic frameworks, as well as clarifying variables and parameters. This stage captures the core modeling choices and reflects whether the model structure is appropriate, coherent, and aligned with the problem formulation and assumptions.

\paragraph{Model Solving.}
Model solving addresses how the constructed model is analyzed or optimized to produce solutions. This includes deriving analytical solutions when possible, selecting numerical methods, specifying optimization procedures, and ensuring solvability. Even a well-designed model may fail at this stage if solution methods are ill-posed, unstable, or computationally infeasible. Separating model solving from model construction allows evaluation to distinguish conceptual modeling quality from solution execution quality.

\paragraph{Code Implementation.}
Code implementation refers to translating the proposed solution procedures into executable code. This stage includes algorithm realization, parameter initialization, data processing, and handling of practical issues such as numerical stability and runtime constraints. Implementation quality determines whether a model can be executed, reproduced, and applied in practice. Evaluating this stage is necessary because conceptual correctness does not guarantee executable or reliable implementation.

\paragraph{Result Analysis.}
Result analysis involves interpreting, validating, and critically examining the outputs produced by the implemented model. This includes checking correctness, assessing robustness, analyzing errors or uncertainty, comparing against baselines or expectations, and discussing limitations. Result analysis closes the modeling loop by determining whether the obtained results meaningfully address the original problem and whether conclusions are supported by evidence.

\subsection{Example of Problem- and Stage-Aware Rubric Instantiation}
\label{app:rubric-example}
\paragraph{Example Subtask.}
We present a concrete example of rubric instantiation for a representative modeling subtask involving \textit{data preprocessing and observation uncertainty modeling}. The subtask requires transforming raw radar observations from multiple sensors into a unified local ENU coordinate system, propagating measurement uncertainty from polar to Cartesian space, handling asynchronous timestamps, and detecting abnormal observations. The following stage-wise criteria illustrate how evaluation is instantiated in a problem- and stage-conditioned manner.

In this example, the abstract dimensions discussed in the main text are instantiated concretely as follows. 
\emph{Modeling objectives} correspond to the explicit preprocessing goals of coordinate unification, uncertainty propagation, and temporal alignment. 
\emph{Modeling assumptions} include geometric conventions, noise distributions, independence hypotheses, and synchronization tolerances. 
\emph{Modeling decisions} are reflected in concrete choices such as coordinate frames, uncertainty representations, linearization strategies, and handling of asynchronous observations. 
\emph{Validation} refers to checks on numerical stability, uncertainty consistency, output completeness, and the impact of preprocessing on downstream tracking and filtering modules.

\paragraph{Stage: Problem Identification.}
Evaluation at this stage focuses on whether the model correctly recognizes the core objectives and requirements implied by the subtask.

\begin{itemize}
  \item \textbf{Core processing objective recognition}: 
  Whether the solution explicitly identifies the need to convert raw polar measurements into a local ENU coordinate system and to produce position estimates accompanied by quantified uncertainty.

  \item \textbf{Observation and uncertainty source identification}: 
  Whether the solution correctly identifies the relevant observation components (range, azimuth, elevation, time, sensor identity) and acknowledges key sources of measurement uncertainty.

  \item \textbf{Downstream interface awareness}: 
  Whether the solution recognizes that the processed outputs must be directly usable by subsequent filtering and data association modules, including requirements on data format, time consistency, and uncertainty representation.
\end{itemize}

\paragraph{Stage: Problem Formulation.}
This stage evaluates whether the identified task is reformulated into an explicit mathematical or algorithmic representation.

\begin{itemize}
  \item \textbf{Measurement mapping specification}: 
  Whether the solution clearly specifies the mathematical mapping from polar measurements to Cartesian coordinates under the given coordinate conventions.

  \item \textbf{Angle convention consistency}: 
  Whether the solution explicitly addresses the transformation between problem-defined angle conventions and standard mathematical function inputs.

  \item \textbf{Uncertainty representation strategy}: 
  Whether the solution articulates how measurement uncertainty is represented and propagated, for example via linearization or by preserving uncertainty in the original measurement space.
\end{itemize}

\paragraph{Stage: Assumption Development.}
Evaluation at this stage focuses on whether simplifying assumptions are explicitly stated and justified.

\begin{itemize}
  \item \textbf{Geometric and physical assumptions}: 
  Whether assumptions regarding Earth model, sensor motion, and coordinate system stability are clearly stated.

  \item \textbf{Noise and independence assumptions}: 
  Whether assumptions about measurement noise distributions and independence are made explicit and their limitations acknowledged.

  \item \textbf{Temporal synchronization assumptions}: 
  Whether assumptions regarding acceptable time misalignment and synchronization strategy across sensors are clearly described.
\end{itemize}

\paragraph{Stage: Model Construction.}
This stage evaluates whether the formulated problem and assumptions are faithfully operationalized.

\begin{itemize}
  \item \textbf{Coordinate transformation implementation}: 
  Whether the solution specifies a complete and consistent transformation pipeline from geographic coordinates to the local ENU frame.

  \item \textbf{Uncertainty propagation realization}: 
  Whether the solution defines how measurement uncertainty is operationalized in the constructed model, including Jacobian-based propagation or alternative representations.

  \item \textbf{Asynchronous observation handling}: 
  Whether the model explicitly incorporates a strategy for handling observations arriving at different timestamps.
\end{itemize}

\paragraph{Stage: Model Solving.}
Evaluation at this stage focuses on whether the constructed model is solved or executed in a numerically and statistically sound manner.

\begin{itemize}
  \item \textbf{Numerical stability considerations}: 
  Whether the solution addresses potential numerical issues arising from extreme angles, near-singular configurations, or linearization errors.

  \item \textbf{Approximation validity checking}: 
  Whether the solution considers the accuracy of linear approximations or validates them through alternative checks or simulations.

  \item \textbf{Time-consistency verification}: 
  Whether the temporal handling of observations produces outputs consistent with the assumed time evolution.
\end{itemize}

\paragraph{Stage: Code Implementation.}
This stage evaluates whether the proposed procedures are realized in an executable and reproducible manner.

\begin{itemize}
  \item \textbf{Executable transformation and propagation modules}: 
  Whether the implementation correctly realizes coordinate transformations and uncertainty propagation as described.

  \item \textbf{Structured output generation}: 
  Whether the implementation produces a structured observation sequence containing positions, uncertainties, timestamps, and sensor metadata.

  \item \textbf{Verification and testing practices}: 
  Whether basic tests or checks are included to verify correctness of transformations, uncertainty propagation, and boundary cases.
\end{itemize}

\paragraph{Stage: Result Analysis.}
Evaluation at this stage focuses on whether the produced outputs are validated and interpreted.

\begin{itemize}
  \item \textbf{Output completeness and consistency}: 
  Whether the resulting observation sequence is complete, temporally ordered, and internally consistent.

  \item \textbf{Uncertainty consistency checking}: 
  Whether the reported uncertainties are examined for statistical consistency or plausibility.

  \item \textbf{Impact on downstream modeling}: 
  Whether the solution discusses how preprocessing choices and uncertainty handling affect subsequent filtering and data association performance.
\end{itemize}

%% file: appendix/data_preprocess.tex
\section{Dataset Preprocessing}
\label{app:data_preprocess}

Many original problem statements include complex mathematical notation and multi-page layouts. To produce a canonical, machine-processable representation while preserving mathematical fidelity, we adopt the following preprocessing pipeline:

\begin{enumerate}
    \item \textbf{Page-level OCR-to-LaTeX conversion:} Each page of a raw problem PDF is converted into LaTeX markup using the multimodal model \texttt{Qwen2.5-VL-32B-Instruct}, instructed to output \LaTeX-compatible representations for inline and display math, numbered equations, tables, and figures. Conversions are performed on a per-page basis to retain layout cues and to avoid cross-page concatenation errors.
    \item \textbf{Stitching and structural normalization:} Page-level \LaTeX fragments are programmatically concatenated into a single problem file. During stitching we normalize equation numbering, caption anchors, cross-references, and enumerations to produce a syntactically consistent \LaTeX source for each problem.
    \item \textbf{Human verification and correction:} Each converted problem undergoes manual review by domain-expert annotators. Verification focuses on: (i) correct transcription of mathematical symbols and subscripts/superscripts, (ii) fidelity of displayed equations and matrices, (iii) correct mapping of figures/tables to captions, and (iv) semantic sanity checks (e.g., unit consistency where applicable). Annotators edit the \LaTeX source as needed to ensure near-perfect fidelity to the original problem statement.
    \item \textbf{Canonicalization and metadata extraction:} We extract and store structured metadata for each problem (domain tags, problem type labels such as ``prediction/optimization/simulation/evaluation'', number of equations, presence of data tables, and estimated modeling complexity) to support downstream stratified analyses.
\end{enumerate}

This pipeline yields a clean, human-verified \LaTeX corpus of graduate modeling problems that preserves symbolic content and is ready for model prompting, archival, and reproducible evaluation.

%% file: appendix/ICC.tex
\section{Intraclass Correlation Coefficient: ICC(2,1)}
\label{app:icc}

\subsection{Motivation and Use Case}

To assess the reliability of human expert judgments in our evaluation framework, we employ the \emph{intraclass correlation coefficient} (ICC)~\cite{weir2005quantifying}, which measures the degree of consistency or conformity among measurements made by multiple observers when assessing the same quantity. For continuous-valued ratings, ICC quantifies the proportion of total score variance that can be attributed to systematic differences among the evaluated items, as opposed to variability arising from individual raters or random measurement noise~\cite{shrout1979intraclass}.

In this work, ICC is used to evaluate the consistency among multiple expert judges when scoring the same set of mathematical modeling outputs. High ICC values indicate that the evaluation scores are stable and reproducible across judges, thereby supporting the validity of the proposed evaluation framework.

\subsection{Model Selection: Why ICC(2,1)}

We adopt ICC(2,1) following the classification of ~\citet{shrout1979intraclass}, which corresponds to a two-way random-effects model with absolute agreement, based on single measurements. This choice is appropriate for our evaluation setting for the following reasons:

\begin{itemize}
    \item \textbf{Two-way design.}  
    Each modeling output is rated by the same group of judges, allowing score variance to be decomposed into item effects, rater effects, and residual error.
    
    \item \textbf{Random-effects assumption for raters.}  
    Judges are treated as a random sample from a broader population of qualified experts. The goal is to assess whether evaluation scores would remain reliable if a different but comparable group of experts were used.
    
    \item \textbf{Absolute agreement.}  
    ICC(2,1) measures absolute agreement rather than rank consistency. This is essential because the evaluation framework produces numerical scores whose absolute values are meaningful.
    
    \item \textbf{Single-measure reliability.}  
    ICC(2,1) evaluates the reliability of a single judge’s score, providing a conservative estimate that aligns with realistic evaluation scenarios.
\end{itemize}

\subsection{Statistical Formulation}

Let \( n \) denote the number of items being rated (e.g., modeling reports), \( k \) the number of raters, and \( Y_{ij} \) the score assigned by rater \( j \) to item \( i \).
The two-way random-effects ANOVA model is defined as:
\[
Y_{ij} = \mu + \alpha_i + \beta_j + \varepsilon_{ij},
\]
where:
\begin{itemize}
    \item \( \mu \) is the grand mean,
    \item \( \alpha_i \sim \mathcal{N}(0, \sigma^2_{\alpha}) \) represents the random effect of item \( i \),
    \item \( \beta_j \sim \mathcal{N}(0, \sigma^2_{\beta}) \) represents the random effect of rater \( j \),
    \item \( \varepsilon_{ij} \sim \mathcal{N}(0, \sigma^2_{\varepsilon}) \) denotes residual error.
\end{itemize}

Based on the corresponding ANOVA mean squares:
\begin{itemize}
    \item \( MS_R \): mean square for items (rows),
    \item \( MS_C \): mean square for raters (columns),
    \item \( MS_E \): residual mean square,
\end{itemize}
the ICC(2,1) is computed as:
\[
\mathrm{ICC}(2,1) =
\frac{MS_R - MS_E}
     {MS_R + (k - 1)MS_E + \frac{k}{n}(MS_C - MS_E)}.
\]

%% file: appendix/case.tex
\section{Diagnostic Case Analysis}
\label{app:case}

This appendix presents three diagnostic case analysis to qualitatively examine the differences between the baseline rubric and our problem-oriented, stage-aware evaluation framework. The goal is to illustrate how high baseline scores may mask critical modeling deficiencies, and how such deficiencies are explicitly identified under our framework.

\subsection{Case Overview}

\begin{table}[ht]
\centering
\begin{adjustbox}{width=\linewidth}
\begin{tabular}{cc|cccc|ccccccc}
\toprule
report-id & subtask-id & Prb Ana & Mod Rig & Pra Sci & Res Bias & Prb Idf & Prb Frm & Asm Dev & Mod Con & Mod Sol & Cod Imp & Res Ays \\
\midrule
\multirow{4}{*}{Case 1} 
& 1 & \multirow{4}{*}{8.5} & \multirow{4}{*}{8.5} & \multirow{4}{*}{9} & \multirow{4}{*}{8.5} & 10 & 9 & 8.5 & 8.5 & 4.3 & 0 & 1.5 \\
& 2 &  &  &  &  & 6.2 & 6.7 & 8.1 & 8.2 & 5.7 & 0 & 0.4 \\
& 3 &  &  &  &  & 8.5 & 8.6 & 8 & 8.9 & 4.8 & 2.5 & 5.6 \\
& 4 &  &  &  &  & 7.8 & 7.1 & 6.9 & 8.3 & 5.2 & 2.8 & 4.2 \\
\hline
\multirow{4}{*}{Case 2} 
& 1 & \multirow{4}{*}{9.5} & \multirow{4}{*}{8.5} & \multirow{4}{*}{9.5} & \multirow{4}{*}{9.5} & 4 & 5.5 & 1.5 & 3 & 2.5 & 0 & 1 \\
& 2 &  &  &  &  & 9 & 8.5 & 4.3 & 0.5 & 3.3 & 1.5 & 3.1 \\
& 3 &  &  &  &  & 8.4 & 9.6 & 6.1 & 8.7 & 5.3 & 5.9 & 2.8 \\
& 4 &  &  &  &  & 9.3 & 9.3 & 6.7 & 9 & 6 & 1.1 & 3 \\
\hline
\multirow{4}{*}{Case 3} 
& 1 & \multirow{4}{*}{9.5} & \multirow{4}{*}{8.5} & \multirow{4}{*}{9} & \multirow{4}{*}{10} & 9.5 & 10 & 9.5 & 6.7 & 3.5 & 1.5 & 1.5 \\
& 2 &  &  &  &  & 6.3 & 4.5 & 8.5 & 8 & 10 & 0.5 & 8 \\
& 3 &  &  &  &  & 10 & 10 & 9.3 & 9.8 & 9.1 & 2 & 8.9 \\
& 4 &  &  &  &  & 9.5 & 9.3 & 7.1 & 9.2 & 8.3 & 3.1 & 8.6 \\
\bottomrule
\end{tabular}
\end{adjustbox}
\caption{Stage-wise evaluation scores for diagnostic cases under the baseline rubric and the proposed problem-conditioned framework. Each case consists of multiple subtasks, illustrating how high baseline scores may coexist with substantial deficiencies at specific modeling stages.  Abbreviations: Prb Ana = Problem Analysis; Mod Rig = Modeling Rigor; Pra Sci = Practicality and Scientificity; Res Bias = Result Bias; Prb Idf = Problem Identification; Prb Frm = Problem Formulation; Asm Dev = Assumption Development; Mod Con = Model Construction; Mod Sol = Model Solving; Cod Imp = Code Implementation; Res Ays = Result Analysis.}

\label{tab:case-overview}
\end{table}

Table~\ref{tab:case-overview} summarizes the selected cases, reporting baseline evaluation scores along the dimensions of problem analysis, modeling rigor, practicality and scientificity, and result bias, together with the corresponding stage-wise scores under our framework.

\subsection{Case~1: High Baseline Score Masking Fundamental Non-Executability}

\paragraph{Baseline Evaluation Behavior.}
Under the baseline rubric, this report receives a high overall evaluation, with scores of 8.5 in problem analysis, modeling rigor, and result bias, and a score of 9.0 in practicality and scientificity (Table~\ref{tab:case-overview}). The solution presents a comprehensive and technically sophisticated modeling pipeline for maneuvering target tracking, covering coordinate transformation, uncertainty propagation, state estimation, data association, and trajectory prediction. Advanced techniques such as IMM filtering, unscented Kalman filtering, probabilistic data association, and Monte Carlo-based prediction are correctly referenced and organized into a seemingly coherent workflow. The report is well structured, uses domain-appropriate terminology, and includes explicit assumptions and detailed mathematical formulations, which are strongly rewarded by the baseline evaluation scheme.

\paragraph{Problem-Grounded Modeling Deficiency.}
Despite its surface-level completeness, the solution fails to satisfy several necessary conditions for valid problem solving. Most critically, key components of the proposed pipeline are never operationalized or verified. Essential elements such as algorithmic parameter specification, statistical gating criteria, and consistency checks are either left unspecified or discussed only at a conceptual level. More importantly, the provided implementation code is not executable and produces no valid outputs. Consequently, none of the proposed modeling choices, including filtering, data association, or trajectory prediction, are empirically validated on the given datasets. Although the high-level modeling intent is reasonable, the solution does not constitute a functioning or verifiable problem-solving pipeline.

\paragraph{Stage-Aware Diagnosis.}
The proposed problem-oriented, stage-aware evaluation framework explicitly exposes these execution-level failures. While early stages such as problem identification and problem formulation consistently receive high scores (typically above 8), later stages exhibit a sharp performance drop. In particular, the code implementation stage receives scores of 0 across multiple subtasks, and result analysis scores fall as low as 0.4, indicating the absence of executable procedures, reproducible outputs, and quantitative validation (Table~\ref{tab:case-overview}). Because the framework enforces stage-specific necessary conditions, these failures cannot be compensated by fluent exposition or sophisticated method descriptions. This case demonstrates how high baseline scores may mask fundamental non-executability, whereas the stage-aware framework reveals the gap between apparent modeling sophistication

\subsection{Case~2: Methodological Sophistication Without Problem Compliance}

\paragraph{Baseline Evaluation Behavior.}
Under the baseline rubric, this report receives consistently high scores across all baseline dimensions, including 9.5 in problem analysis, 8.5 in modeling rigor, 9.5 in practicality and scientificity, and 9.5 in result bias (Table~\ref{tab:case-overview}). The solution presents a technically rich and well-articulated modeling pipeline for molecular property prediction, covering feature preprocessing, regression modeling, classification, uncertainty quantification, multi-objective screening, and interpretability analysis. Advanced techniques such as sparse group PLS, Gaussian process residual modeling, nested cross-validation, SMOTE-based class balancing, probabilistic calibration, Pareto optimization, and SHAP-based explanation are systematically introduced and mathematically formalized. The report is comprehensive, logically structured, and demonstrates strong familiarity with modern cheminformatics and machine learning practices, which aligns well with the surface-level criteria emphasized by the baseline evaluation.

\paragraph{Problem-Grounded Modeling Deficiency.}
Despite its methodological sophistication, the solution fails to satisfy key problem-specific modeling requirements. In multiple subtasks, the implemented procedures deviate substantially from the methods explicitly specified in the problem description. For example, fixed-threshold rules mandated by the task, such as missing-value and variance filtering criteria, correlation-based redundancy removal, and prescribed PLS--VIP--LASSO pipelines, are replaced by alternative strategies including adaptive quantile thresholds, hierarchical clustering, sparse group regularization, and hybrid models. While these alternatives are individually reasonable, they are not justified as equivalent to the required procedures, nor is their impact on task objectives analyzed. As a result, the solution effectively addresses a different modeling problem than the one posed, without verifying compliance with the original task constraints and acceptance criteria.

\paragraph{Stage-Aware Diagnosis.}
The stage-aware evaluation framework explicitly exposes this form of problem non-compliance. Although early stages such as problem identification and problem formulation often receive moderate to high scores (typically above 8 in several subtasks), sharp performance drops are observed in assumption development, model construction, and model solving. In particular, assumption development scores fall as low as 1.5, model construction scores drop to 0.5 in some subtasks, and model solving scores remain in the low range of approximately 2.5–3.3, reflecting unverified deviations from task-specified procedures (Table~\ref{tab:case-overview}). Later stages further exhibit weak performance in code implementation and result analysis, with code implementation scores reaching 0 and result analysis scores as low as 1.0. By enforcing problem-conditioned, stage-specific criteria, the framework distinguishes methodological sophistication from task-faithful problem solving. This case demonstrates how baseline evaluation may reward technically advanced but misaligned solutions, whereas the proposed framework identifies failures arising from solving an unintended problem.

\subsection{Case~3: Scientific Rigor Without Task-Level Executability}

\paragraph{Baseline Evaluation Behavior.}
Under the baseline rubric, this report receives near-ceiling scores across all baseline dimensions, including 9.5 in problem analysis, 8.5 in modeling rigor, 9.0 in practicality and scientificity, and a perfect score of 10 in result bias (Table~\ref{tab:case-overview}). The solution presents an exceptionally rigorous statistical modeling framework for longitudinal edema progression analysis, incorporating generalized additive mixed models, functional principal component analysis, clustering-based subgroup discovery, causal inference via propensity score weighting and marginal structural models, and joint modeling of hematoma--edema dynamics. Model formulations are mathematically precise, assumptions are explicitly discussed, and results are interpreted with strong domain knowledge and clinical insight. From a baseline perspective, the report closely resembles a polished scientific study and is therefore strongly rewarded.

\paragraph{Problem-Grounded Modeling Deficiency.}
Despite its high scientific quality, the solution fails to fully satisfy the task-specific execution requirements of the modeling problem. Several required deliverables are never concretely produced or verified. Although the report specifies detailed modeling pipelines and estimation procedures, it does not provide executable implementations, reproducible scripts, or concrete intermediate artifacts such as constructed long-format datasets, time-mapped measurement tables, or quality control logs. Moreover, key task-mandated outputs, including patient-level subgroup assignments, numerical result tables, and verifiable intermediate results, are either described only conceptually or reported in aggregated form. As a result, the solution remains at the level of a methodological proposal and empirical summary rather than a task-complete, checkable problem solution.

\paragraph{Stage-Aware Diagnosis.}
The stage-aware evaluation framework exposes this gap between analytical rigor and task-level execution. While early stages such as problem identification, problem formulation, and assumption development consistently receive very high scores (often above 9), later stages exhibit a pronounced performance drop. In particular, code implementation scores range from 1.5 to 3.1 across subtasks, and result analysis scores fall as low as 1.5, reflecting the absence of executable pipelines, reproducible artifacts, and task-specified outputs (Table~\ref{tab:case-overview}). By enforcing problem-conditioned and stage-specific criteria, the framework prevents scientific sophistication from compensating for missing execution and verification. This case illustrates how baseline evaluation may reward analytically impressive but operationally incomplete solutions, whereas the proposed framework reliably identifies failures in task-level completion.

%% file: appendix/stage_faliure_causes.tex
\section{Operational Definitions of Failure Dimensions}
\label{app:failure-causes}

\begin{longtable}{p{2.8cm} p{3.5cm} p{8cm}}
\caption{Operational Definitions of Stage-Specific Failure Causes Corresponding to Table~\ref{tab:failure-causes}} \\
\toprule
Stage & Failure Cause & Operational Definition \\
\midrule
\endfirsthead

\toprule
Stage & Failure Cause & Operational Definition \\
\midrule
\endhead

\bottomrule
\endfoot

\bottomrule
\endlastfoot

Assumption Development
& Assumptions Not Checked
& Assumptions that are in principle testable are stated but no validation, diagnostic procedure, or empirical check is provided. \\

Assumption Development
& Missing Assumption Conditions
& Necessary conditions under which an assumption holds (e.g., constraints, statistical tests, parameter bounds) are not explicitly specified. \\

Assumption Development
& Assumption Impact Not Discussed
& The influence of assumptions on model behavior, results, or conclusions is not analyzed or discussed. \\

Assumption Development
& Hidden Idealized Assumptions
& Idealized premises (e.g., noise-free data, full observability, stationarity) are implicitly adopted without being explicitly declared. \\

Assumption Development
& Unrealistic Assumptions
& Assumptions are incompatible with the problem context, data-generating process, or domain knowledge, even if explicitly stated. \\

Model Construction
& Incomplete Model Structure
& The model lacks essential components such as key variables, constraints, objective functions, or state relationships required to represent the task. \\

Model Construction
& Missing Model Derivation
& The transition from problem description to formal model lacks sufficient derivation steps or logical justification. \\

Model Construction
& Unclear Variables or Symbols
& Variables, symbols, or dimensional definitions are ambiguous, inconsistent, or insufficiently specified. \\

Model Construction
& Solvability Not Justified
& The mathematical feasibility, identifiability, or solvability of the model is not established or argued. \\

Model Construction
& Model Deviates from Task Goal
& The chosen model formulation or abstraction fails to align with the stated modeling objectives or task requirements. \\

Model Construction
& Model--Assumption Conflict
& The formal model contradicts, ignores, or fails to operationalize the stated assumptions. \\

Model Solving
& No Checkable Solution
& No numerical solution, concrete output, or verifiable result is provided for the constructed model. \\

Model Solving
& Key Solution Steps Missing
& Critical derivation steps, algorithmic procedures, or implementation details required to reproduce the solution are absent. \\

Model Solving
& Solution Not Verified
& The correctness, feasibility, convergence, or optimality of the solution is not validated or demonstrated. \\

Model Solving
& Numerical Stability Not Analyzed
& Sensitivity, robustness, or numerical stability with respect to parameters or numerical settings is not examined. \\

Model Solving
& Computationally Infeasible
& The proposed solution approach is computationally impractical at the intended scale, without mitigation or discussion. \\

Model Solving
& Inappropriate Solution Method
& The solution method or algorithm is theoretically unsuitable for the given model structure. \\

Code Implementation
& Cannot Be Executed
& The provided code or implementation cannot be executed due to errors, missing components, or invalid configurations. \\

Code Implementation
& Results Not Reproducible
& The implementation lacks sufficient information (e.g., parameters, seeds, environment details) to reproduce reported results. \\

Code Implementation
& No Code
& No executable code, pseudocode, or inspectable implementation is provided. \\

Code Implementation
& Engineering or Numerical Risks
& The implementation exhibits serious issues related to numerical stability, efficiency, resource usage, or error handling. \\

Code Implementation
& Code--Model Mismatch
& The implementation does not faithfully realize the mathematical model or solution logic. \\

Result Analysis
& No Meaningful Results
& No analyzable numerical results, experimental outputs, or substantive analysis targets are presented. \\

Result Analysis
& No Validation or Comparison
& No baseline comparison, validation experiment, or reference analysis is conducted to contextualize results. \\

Result Analysis
& Sensitivity Not Analyzed
& Dependence of results on key assumptions, parameters, or inputs is not examined. \\

Result Analysis
& Conclusions Not Supported
& The reported results are insufficient to justify the stated conclusions. \\

Result Analysis
& Results Miss the Goal
& The analysis fails to address the original modeling objectives or evaluation criteria. \\

Result Analysis
& Limits Not Discussed
& The scope of applicability, failure conditions, or limitations of the results are not discussed. \\

\end{longtable}

%% file: appendix/RQ3_supplementary.tex
\section{Supplementary Analyses for RQ3}
\label{app:rq3_supple}

To better understand the reasons underlying low scores, we conduct a stage-wise analysis of failure causes across the full mathematical modeling workflow. For the \emph{assumption development}, \emph{model construction}, \emph{model solving}, \emph{code implementation}, and \emph{result analysis} stages, we collect all reviewer comments associated with subtasks whose stage-level scores fall below a fixed threshold (8 points). These comments are then summarized and grouped according to the recurring issues they explicitly identify, yielding a set of representative failure causes for each stage.
Table~\ref{tab:failure-causes} provides an overview of the resulting failure causes identified at each stage.

\subsection{Assumption Development}
\label{subsec:assumption-development}

\begin{table*}[ht]
\centering
\small
\begin{tabular}{l c c c c c c}
\toprule
Model  & Not Checked & Miss Cond. & Impact Ignored & Hidden Ideal. & Unrealistic \\
\midrule
Qwen3-235B  & 0.7000 & 0.5625 & 0.4250 & 0.4125 & 0.0000 \\
DS-Inst    & 0.6743 & 0.5829 & 0.3771 & 0.4629 & 0.0057 \\
DS-Think   & 0.6872 & 0.6089 & 0.4525 & 0.3799 & 0.0056 \\
o4-mini    & 0.7412 & 0.6518 & 0.4441 & 0.4409 & 0.0128 \\
\bottomrule
\end{tabular}
\caption{
Distribution of primary failure causes in the assumption development stage.
Model abbreviations: Qwen3-235B = Qwen3-235B-A22B-Instruct-2507;
DS-Inst = DeepSeek-V3.2-Instruct; DS-Think = DeepSeek-V3.2-Thinking.
Failure causes denote:
\emph{Not Checked} (Assumptions Not Checked),
\emph{Miss Cond.} (Missing Assumption Conditions),
\emph{Impact Ignored} (Assumption Impact Not Discussed),
\emph{Hidden Ideal.} (Hidden Idealized Assumptions),
and \emph{Unrealistic} (Unrealistic Assumptions).
\emph{Total} denotes the number of subtasks whose assumption-stage score is below~8.
}
\label{tab:assumption-failure-causes}
\end{table*}

Table~\ref{tab:assumption-failure-causes} shows that low scores in the assumption development stage are primarily driven by execution-level shortcomings rather than fundamentally flawed assumptions.
Across all models, \emph{Not Checked} is by far the most prevalent failure cause, accounting for approximately 78--83\% of low-scoring cases.
This indicates that models frequently introduce assumptions that are, in principle, verifiable, yet fail to provide any accompanying tests, diagnostics, or concrete validation procedures.

The second most common issue is \emph{Impact Ignored}, with frequencies ranging from roughly 42\% to 66\%.
This suggests that even when assumptions are stated, models often do not analyze how these assumptions affect downstream modeling results or conclusions, nor do they discuss the conditions under which the assumptions may break down.

\emph{Missing Conditions} also appears consistently across models (around 35--42\%), revealing a systematic tendency to state assumptions without clearly specifying the prerequisites under which they hold, such as required statistical properties, stability conditions, or parameter constraints.
In contrast, \emph{Idealized} assumptions occur at a moderate but non-negligible rate (approximately 11--17\%), indicating that models sometimes rely on implicit idealized premises without explicitly acknowledging them.

Finally, \emph{Unrealistic} assumptions are relatively rare, remaining below 6\% for all models.
This suggests that models generally avoid making assumptions that are obviously incompatible with the problem setting or underlying mechanisms.

Overall, failures in the assumption development stage are dominated by insufficient operationalization and reflection—namely, a lack of validation, condition specification, and impact analysis—rather than by the introduction of fundamentally unreasonable assumptions.

\subsection{Model Construction}
\label{subsec:model-construction}

\begin{table*}[ht]
\centering
\small
\begin{tabular}{l c c c c c c c}
\toprule
Model  & Incomp. Struct. & Var. Unclear & Miss Deriv. & Solv. Unjust. & Goal Deviat. & Assump. Conflict \\
\midrule
DS-Inst    & 0.6604 & 0.4717 & 0.2075 & 0.0377 & 0.0377 & 0.0566 \\
DS-Think    & 0.8030 & 0.3182 & 0.0909 & 0.0303 & 0.1364 & 0.0303 \\
Qwen3-235B & 0.8125 & 0.3750 & 0.1250 & 0.0625 & 0.0625 & 0.0000 \\
o4-mini   & 0.4560 & 0.6080 & 0.1840 & 0.1920 & 0.0320 & 0.0240 \\
\bottomrule
\end{tabular}
\caption{
Distribution of primary failure causes in the model construction stage.
Model abbreviations: DS-Inst = DeepSeek-V3.2-Instruct; 
DS-Think = DeepSeek-V3.2-Thinking; 
Qwen3-235B = Qwen3-235B-A22B-Instruct-2507.
Failure causes include:
\emph{Miss Deriv.} (Missing Model Derivation),
\emph{Incomp. Struct.} (Incomplete Model Structure),
\emph{Goal Deviat.} (Model Deviates from Task Goal),
\emph{Var. Unclear} (Unclear Variables or Symbols),
\emph{Solv. Unjust.} (Solvability Not Justified),
and \emph{Assump. Conflict} (Model--Assumption Conflict).
\emph{Total} denotes the number of subtasks whose model-construction score is below~8.
f}
\label{tab:model-construction-failure-causes}
\end{table*}

As shown in Table~\ref{tab:model-construction-failure-causes}, low scores in the model construction stage are dominated by deficiencies in how models are formulated and justified, rather than by overtly incorrect modeling choices.
The most prevalent failure cause is \emph{Lack Derivation}, which accounts for more than half of the low-scoring cases across all models.
This indicates that models often present formal equations or structural forms without clearly articulating the intermediate reasoning steps that connect the problem description to the proposed model.

Another major contributor is \emph{Incomplete}, with proportions ranging from approximately 48\% to 69\%.
This suggests that essential components of the model, such as key variables, constraints, or objective terms, are frequently missing or only partially specified, resulting in models that are structurally insufficient even if their overall direction is plausible.

By contrast, failures related to \emph{Goal Mismatch}, \emph{Variables}, \emph{Solvability}, and \emph{Assump.\ Conflict} occur at substantially lower rates.
This implies that, once a model structure is proposed, it is usually aligned with the task objective and internally coherent, but often lacks completeness or adequate justification.

Overall, these findings indicate that low scores in the model construction stage are primarily driven by underspecified and weakly motivated formulations, rather than by fundamentally misguided modeling decisions.
For completeness, Appendix~\ref{app:failure-causes} provides formal definitions of each failure cause used in this stage, together with representative reviewer comment excerpts that illustrate how these issues manifest in practice.

\subsection{Model Solving}
\label{subsec:model-solving}

\begin{table*}[ht]
\centering
\small
\begin{tabular}{l c c c c c c c}
\toprule
Model  & No Output  & Steps Missing  & Not Verified  & No Stability  & Infeasible  & Wrong Method \\
\midrule
DS-Inst     & 0.3592 & 0.4296 & 0.3099 & 0.3310 & 0.0704 & 0.0070 \\
DS-Think    & 0.3896 & 0.4481 & 0.3506 & 0.2727 & 0.0844 & 0.0260 \\
Qwen3-235B   & 0.5287 & 0.3333 & 0.2414 & 0.1954 & 0.0345 & 0.0115 \\
o4-mini    & 0.5855 & 0.2566 & 0.5263 & 0.2730 & 0.0658 & 0.0197 \\
\bottomrule
\end{tabular}
\caption{
Distribution of primary failure causes in the model solving stage.
Model abbreviations: DS-Inst = DeepSeek-V3.2-Instruct; 
DS-Think = DeepSeek-V3.2-Thinking; 
Qwen3-235B = Qwen3-235B-A22B-Instruct-2507.
Failure causes include:
\emph{No Output} (No Checkable Solution),
\emph{Steps Missing} (Key Solution Steps Missing),
\emph{Not Verified} (Solution Not Verified),
\emph{No Stability} (Numerical Stability Not Analyzed),
\emph{Infeasible} (Computationally Infeasible),
and \emph{Wrong Method} (Inappropriate Solution Method).
\emph{Total} denotes the number of subtasks whose model-solving score is below~8.
}
\label{tab:model-solving-failure-causes}
\end{table*}

As shown in Table~\ref{tab:model-solving-failure-causes}, low scores in the model solving stage are primarily driven by deficiencies in solution verification and procedural completeness, rather than by the absence of solution attempts.
Across all models, \emph{Unverified} is the most prevalent failure cause, reaching up to 72\% for some models.
This indicates that although solutions are often proposed, their feasibility, optimality, or convergence is rarely demonstrated or justified.

A second major contributor is \emph{Missing Steps}, which appears in approximately 30--47\% of low-scoring cases.
This suggests that solution procedures are frequently described only at a high level, omitting critical intermediate steps needed to assess correctness or ensure reproducibility.

Failures related to \emph{No Stability} are also non-negligible, reflecting a systematic lack of sensitivity analysis or numerical stability assessment, even when such analyses are necessary for validating the robustness of the proposed solution.
By contrast, \emph{Invalid Method}, \emph{No Output}, and \emph{Infeasible} occur at substantially lower rates.
This implies that models generally avoid proposing solutions that are outright inappropriate, absent, or computationally impractical, but often fail to rigorously justify the solutions they do present.

Overall, low scores in the model solving stage arise not from missing solutions, but from insufficient verification, incomplete procedural exposition, and limited robustness analysis.
For clarity and reproducibility, Appendix~\ref{app:failure-causes} provides formal definitions of each failure cause used in this stage, together with representative reviewer comment excerpts that illustrate how these issues manifest in practice.

\subsection{Code Implementation}
\label{subsec:code-implementation}
\begin{table*}[ht]
\centering
\small
\begin{tabular}{l c c c c c c}
\toprule
Model & Results N/R & No Code & Miss Comp. & Eng./Num. Risk & Code--Model Mis. \\
\midrule
DS-Inst     & 0.7564 & 0.5782 & 0.1564 & 0.0800 & 0.0036 \\
DS-Think   & 0.7600 & 0.6133 & 0.2100 & 0.0933 & 0.0000 \\
Qwen3-235B & 0.7238 & 0.6000 & 0.2333 & 0.1143 & 0.0048 \\
o4-mini    & 0.7646 & 0.8915 & 0.3175 & 0.0556 & 0.0026 \\
\bottomrule
\end{tabular}
\caption{
Distribution of primary failure causes in the code implementation stage.
Model abbreviations: DS-Inst = DeepSeek-V3.2-Instruct; 
DS-Think = DeepSeek-V3.2-Thinking; 
Qwen3-235B = Qwen3-235B-A22B-Instruct-2507.
Failure causes include:
\emph{Results N/R} (Results Not Reproducible),
\emph{No Code} (No Usable Code),
\emph{Miss Comp.} (Missing Key Code Components),
\emph{Eng./Num. Risk} (Engineering or Numerical Risks),
and \emph{Code--Model Mis.} (Code--Model Mismatch).
\emph{Total} denotes the number of subtasks whose code-implementation score is below~8.
}
\label{tab:code-implementation-failure-causes}
\end{table*}

Table~\ref{tab:code-implementation-failure-causes} indicates that low scores in the code implementation stage are overwhelmingly driven by reproducibility and executability issues, rather than by discrepancies between the implementation and the underlying model.
Across all models, the most dominant failure cause is \emph{Results Not Reprod.}, accounting for approximately 74--79\% of low-scoring cases.
This suggests that, even when code is provided, the reported results cannot be reliably reproduced due to missing configuration details, unstable execution behavior, or insufficient documentation of experimental settings.

A second major contributor is \emph{No Executable}, which appears at particularly high rates for some models.
This reflects a frequent absence of inspectable or runnable implementations, indicating that solutions are often described at a conceptual or pseudo-code level without providing executable artifacts that allow verification.
In addition, \emph{Missing Modules} occurs consistently across models (around 29--38\%), revealing that critical components of the implementation pipeline, such as key functions, data processing steps, or parameter initialization routines, are often omitted or left unspecified.

By contrast, failures related to \emph{Engineering Risks} appear at moderate levels, suggesting occasional issues such as numerical instability or unsafe engineering practices.
Meanwhile, \emph{Model Mismatch} and \emph{Code Not Reprod.} occur at very low rates, indicating that implementations are generally aligned with the intended model structure and that the code itself is usually provided in a consistent form.
Taken together, these results imply that low scores in the implementation stage stem primarily from insufficient support for execution and reproducibility, rather than from fundamental errors in translating models into code.

For completeness and transparency, Appendix~\ref{app:failure-causes} provides formal definitions of each failure cause used in this stage, along with representative reviewer comment excerpts that illustrate how these implementation-level issues arise in practice.

\subsection{Result Analysis}
\label{subsec:result-analysis}

\begin{table*}[ht]
\centering
\small
\begin{tabular}{l c c c c c c c}
\toprule
Model & No Valid. & No Results & No Sensit. & Weak Concl. & Goal Miss. & No Limits \\
\midrule
DS-Inst   & 0.5417 & 0.4722 & 0.3565 & 0.2639 & 0.0417 & 0.0463 \\
DS-Think   & 0.6161 & 0.4420 & 0.3839 & 0.2991 & 0.0536 & 0.0313 \\
Qwen3-235B& 0.6182 & 0.5818 & 0.3333 & 0.2667 & 0.0061 & 0.0364 \\
o4-mini  & 0.6183 & 0.7984 & 0.3495 & 0.2957 & 0.0323 & 0.0349 \\
\bottomrule
\end{tabular}
\caption{
Distribution of primary failure causes in the result analysis stage.
Model abbreviations: DS-Inst = DeepSeek-V3.2-Instruct; 
DS-Think = DeepSeek-V3.2-Thinking; 
Qwen3-235B = Qwen3-235B-A22B-Instruct-2507.
Failure causes include:
\emph{No Valid.} (No Validation or Comparison),
\emph{No Results} (No Meaningful Results),
\emph{No Sensit.} (Sensitivity Not Analyzed),
\emph{Weak Concl.} (Conclusions Not Supported),
\emph{Goal Miss.} (Results Miss the Goal),
and \emph{No Limits} (Limits Not Discussed).
\emph{Total} denotes the number of subtasks whose result-analysis score is below~8.
}
\label{tab:result-analysis-failure-causes}
\end{table*}

As shown in Table~\ref{tab:result-analysis-failure-causes}, low scores in the result analysis stage are predominantly driven by insufficient empirical validation rather than by isolated presentation errors.
Across all models, \emph{No Validation} is the most frequent failure cause, exceeding 65\% in every case.
This indicates that results are often reported without accompanying baselines, comparative evaluations, or validation experiments that would substantiate the claimed conclusions.

A second prominent issue is \emph{Weak Evidence}, with substantial proportions across models.
This suggests that conclusions are frequently overstated relative to the strength or completeness of the reported results.
In addition, \emph{No Results} remains non-negligible, particularly for some models, reflecting cases where outputs are either absent or insufficiently structured to support meaningful analysis.

Failures related to \emph{No Sensitivity} and \emph{Goal Misaligned} occur at moderate levels, indicating that models often do not analyze how results depend on assumptions or parameters, nor do they consistently align reported outcomes with the stated modeling objectives.
By contrast, \emph{No Limits} appears infrequently, implying that most submissions do not progress to a stage where applicability boundaries or failure conditions are explicitly discussed.

Overall, low scores in the result analysis stage primarily reflect a disconnect between reported outcomes and rigorous empirical grounding, rather than a lack of attempted analysis.
For clarity and reproducibility, Appendix~\ref{app:failure-causes} provides formal definitions of each failure cause used in this stage, along with representative reviewer comment excerpts illustrating how these issues manifest in practice.

\subsection{Summary of Stage-wise Failure Characteristics}
\label{subsec:stage-wise-summary}

Across stages, a consistent pattern emerges: low scores are rarely caused by fundamentally invalid ideas, but instead by insufficient specification, justification, validation, and reproducibility. Early-stage failures tend to involve underspecified assumptions and incomplete model derivations, while later-stage failures increasingly reflect missing verification, weak empirical support, and reproducibility issues. This progression suggests that errors propagate through the modeling pipeline, accumulating rather than being corrected in subsequent stages.

%% file: appendix/problem_analysis.tex
\section{Distribution of China Postgraduate Mathematical Contest in Modeling  Problems }
\label{app:problem-analysis}

\begin{figure}[ht]
  \centering
  \includegraphics[width=0.5\linewidth]{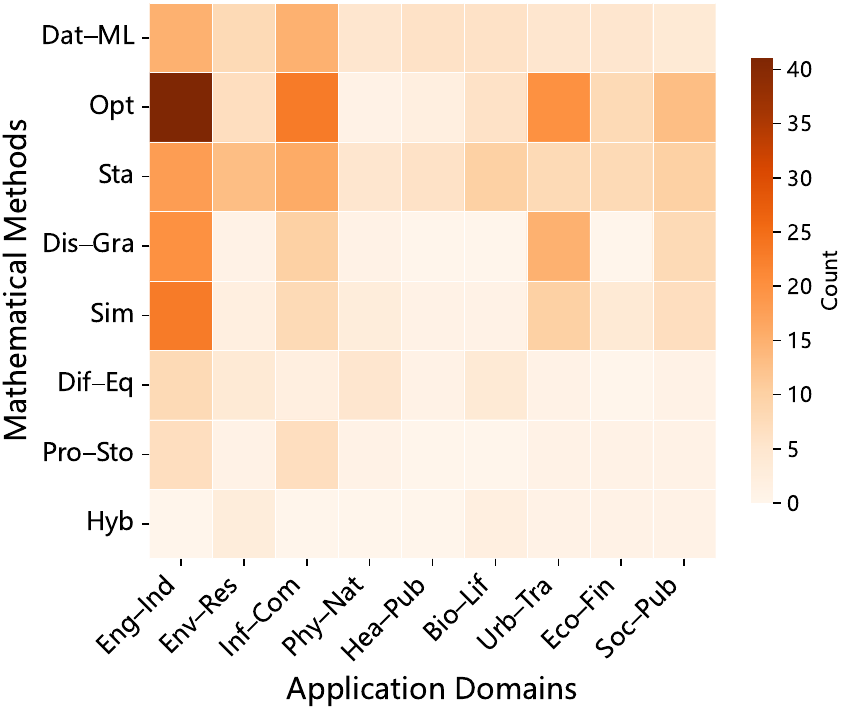}
  \caption{Heatmap of co-occurrence between mathematical method categories and application domains.
  Color intensity indicates the number of modeling problems associated with each method--domain pair.
  Each problem may belong to multiple mathematical methods and application domains; counts therefore reflect multi-label co-occurrences.
  Abbreviations are used for compact visualization: Dat--ML (Data-driven and Machine Learning), Opt (Optimization), Sta (Statistical), Dis--Gra (Discrete and Graph-based), Sim (Simulation-based), Dif--Eq (Differential Equation), Pro--Sto (Probabilistic and Stochastic), Hyb (Hybrid);
  Eng--Ind (Engineering and Industrial), Env--Res (Environmental and Resource), Inf--Com (Information and Communication), Phy--Nat (Physical and Natural), Hea--Pub (Healthcare and Public Health), Bio--Lif (Biological and Life), Urb--Tra (Urban and Transportation), Eco--Fin (Economic and Financial), and Soc--Pub (Social and Public).}
  \label{fig:method-domain-heatmap}
\end{figure}

The dataset consists of 97 problems drawn from the China Postgraduate Mathematical Contest in Modeling (China PMCM), which cover a wide range of real-world modeling scenarios.
To provide a descriptive overview of the modeling tasks used in our evaluation, we analyze their distribution across predefined mathematical method categories and application domains. The \emph{mathematical method categories} describe the primary modeling techniques and mathematical structures required to formulate and solve a problem.
Specifically, we consider differential equation models, optimization models, statistical models, probabilistic and stochastic models, discrete and graph-based models, simulation-based models, data-driven and machine learning models, as well as hybrid modeling approaches that combine multiple methods. The \emph{application domains characterize} the real-world systems or contexts from which the modeling problems arise.
We adopt a coarse-grained domain taxonomy that includes physical and natural systems, engineering and industrial systems, information and communication systems, biological and life systems, healthcare and public health systems, environmental and resource systems, economic and financial systems, social and public systems, and urban and transportation systems.

Each modeling problem is annotated from these two complementary perspectives.
Because mathematical modeling problems are inherently multi-faceted, a single problem may involve multiple mathematical methods and span multiple application domains.
Figure~\ref{fig:method-domain-heatmap} visualizes the resulting co-occurrence statistics using a heatmap representation.
Rather than enforcing exclusive assignments, we adopt a multi-label annotation scheme to reflect the integrated nature of real-world modeling tasks.
Accordingly, the reported counts indicate co-occurrence frequencies between method and domain categories and should be interpreted as descriptive statistics rather than mutually exclusive frequencies.
This analysis is intended to contextualize the diversity of the evaluation tasks and does not imply any causal or performance-related relationships.
The distribution further shows that problems associated with engineering and industrial systems account for a substantial proportion of the dataset.
This reflects the practical orientation of the China PMCM, where many tasks are grounded in real-world engineering scenarios involving system design, resource allocation, process optimization, and operational decision-making. 

Correspondingly, optimization models and statistical models appear with relatively high frequency among the mathematical method categories.
Optimization models are commonly employed to formalize decision-making objectives under practical constraints, which are prevalent in engineering and industrial contexts.
Statistical models, in turn, are frequently used to handle uncertainty, variability, and data-driven inference arising from real-world measurements and observations.
These patterns are consistent with the applied nature of the competition problems and provide contextual insight into the methodological and domain diversity of the evaluation tasks, rather than indicating any methodological preference or performance implication.

%% file: appendix/prompts/criteria.tex
\section{Criteria Generation}
\begin{tcolorbox}[
colback=purple!5!white,
colframe=purple!75!black,
title=Evaluation Criterias Prompt Design,
fonttitle=\bfseries,
boxrule=0.5mm,
arc=2mm,
left=2mm,
right=2mm,
top=2mm,
bottom=2mm,
breakable
]

背景信息：

\{\{background\}\}

问题描述：

\{\{question\}\}

\vspace{3mm}
你现在的任务是：针对一个已明确的数学建模子任务，生成
\textbf{基于任务语义理解的细粒度评分细则}。

\vspace{3mm}
\{\% if previous\_subtasks \%\}

依赖子任务：
\{\{previous\_subtasks\}\}

\{\% endif \%\}

当前子任务：

\{\{subtask\}\}

\vspace{3mm}
—— 任务说明（请严格遵守） ——

你的任务是：针对上面已明确的
\textbf{单一数学建模子任务}，
生成
\textbf{基于任务语义理解的细粒度评分细则（JSON）}，
具体要求如下。

\vspace{3mm}
一、先进行
\textbf{任务语义理解与抽象}
（输出项位于 \texttt{task\_understanding}）：

\begin{enumerate}

\item \texttt{core\_goal}：一句话概括该子任务的 \textbf{核心建模目标}，必须包含子任务中的关键动作词（如“预测”“优化”“拟合”“评估”）和对象（如“AQI”“功率曲线”“成本”）。

\item \texttt{expected\_output}：明确列出预期输出的形式（数值序列、最优解、图表、报表、模型文件、代码等）。

\item \texttt{key\_inputs\_constraints}：列出\textbf{具体输入变量名 / 数据类型 / 必要前置条件 / 关键约束}（至少 3 项；若不足请写“无”并说明原因）。

\item  \texttt{modeling\_type}：从\texttt{\{预测/优化/仿真/拟合/分类/评价/混合\}}中选择最贴切的一项（可复合，如“预测+优化”）。

\item  \texttt{role\_in\_pipeline}：说明该子任务在整体建模流程中的定位，并明确其直接依赖的前序任务与后续任务（例如“作为特征工程后、模型训练前的子任务”）。
\end{enumerate}

\vspace{3mm}
二、基于上述 \texttt{task\_understanding}，
为以下
\textbf{7 个维度}
生成细化评分细则（输出项位于 \texttt{evaluation\_criteria}）：

\begin{itemize}
\item 维度列表（顺序不可更改）：问题识别、问题复述、假设建立、模型构建、模型求解、代码实现、结果分析。

\item 每个维度必须包含 \textbf{3--5 个具体 sub\_criteria}，且每个 sub\_criteria 必须包含以下字段：
    \begin{itemize}
      \item \texttt{sub\_criteria}：评分点主题（必须包含任务关键词）
      \item \texttt{description}：与该子任务语义深度绑定的评分说明
      \item \texttt{score}：该评分点在该维度中的整数分值 （同一维度内总分必须等于 100）
      \item \texttt{evaluation\_focus}：考察能力标签（不超过 6 个词）
      \item \texttt{scoring\_hint}：区分高分 / 中分 / 低分的可判定标准
    \end{itemize}

\item 评分点撰写规则：
    \begin{enumerate}
      \item 必须直接使用 \texttt{task\_understanding} 中的任务关键词。
      \item 禁止使用泛化或空泛表达（如“目标是否明确”）。
      \item \texttt{scoring\_hint} 应给出可操作、可检验的判定标准；
      优先使用量化指标。
      \item 若某维度不适用，
      该维度应返回空数组，
      并增加 \texttt{not\_applicable\_reason} 说明原因。
    \end{enumerate}

\end{itemize}

\vspace{3mm}
三、输出格式与其它要求：

\begin{itemize}
  \item 只返回一个合法的 JSON 对象，不得输出任何多余文字
  \item JSON 编码必须为 UTF-8
  \item 字段顺序必须为 （\texttt{task\_understanding} 在前， \texttt{evaluation\_criteria} 在后）
\end{itemize}

\vspace{3mm}
输出格式示例如下：

\begin{verbatim}
```json
{
  "task_understanding": {
    "core_goal": "...",
    "expected_output": "...",
    "key_inputs_constraints": "...",
    "modeling_type": "...",
    "role_in_pipeline": "..."
  },
  "evaluation_criteria": {
    "问题识别": [],
    "问题复述": [],
    "假设建立": [],
    "模型构建": [],
    "模型求解": [],
    "代码实现": [],
    "结果分析": []
  }
}
```
\end{verbatim}
\end{tcolorbox}

\begin{tcolorbox}[
colback=orange!5!white,
colframe=orange!75!black,
title=Evaluation Criteria Prompt Design,
fonttitle=\bfseries,
boxrule=0.5mm,
arc=2mm,
left=2mm,
right=2mm,
top=2mm,
bottom=2mm,
breakable
]

Background Information:

\{\{background\}\}

Problem Description:

\{\{question\}\}

\vspace{3mm}
Your task is to generate
\textbf{fine-grained evaluation criteria grounded in task-level semantic understanding}
for a clearly specified mathematical modeling subtask.

\vspace{3mm}
\{\% if previous\_subtasks \%\}

Dependent Subtasks:
\{\{previous\_subtasks\}\}

\{\% endif \%\}

Current Subtask:

\{\{subtask\}\}

\vspace{3mm}
—— Task Instructions (Must Be Strictly Followed) ——

Your task is to generate
\textbf{fine-grained, task-aware evaluation criteria (in JSON format)}
for the above specified
\textbf{single mathematical modeling subtask},
according to the requirements below.

\vspace{3mm}
\textbf{I. Task Semantic Understanding and Abstraction}  
(Output under the field \texttt{task\_understanding}):

\begin{enumerate}

\item \texttt{core\_goal}:  
Summarize the \textbf{core modeling objective} of the subtask in one sentence.
The description must include key action verbs (e.g., ``predict'', ``optimize'', ``fit'', ``evaluate'')
and task-specific objects (e.g., ``AQI'', ``power curve'', ``cost'').

\item \texttt{expected\_output}:  
Clearly specify the expected form of the output
(e.g., numerical sequences, optimal solutions, plots, reports, model files, or code).

\item \texttt{key\_inputs\_constraints}:  
List the \textbf{concrete input variables / data types / required preconditions / key constraints}
(at least three items; if fewer, explicitly state ``none'' and explain why).

\item \texttt{modeling\_type}:  
Select the most appropriate type from
\texttt{\{prediction/optimization/simulation/fitting/classification/evaluation/hybrid\}}.
Composite types (e.g., ``prediction+optimization'') are allowed.

\item \texttt{role\_in\_pipeline}:  
Describe the role of this subtask within the overall modeling pipeline,
and explicitly state its direct upstream and downstream dependencies
(e.g., ``performed after feature engineering and before model training'').

\end{enumerate}

\vspace{3mm}
\textbf{II. Fine-Grained Evaluation Criteria Construction}  

Based on the above \texttt{task\_understanding},
generate evaluation criteria for the following
\textbf{seven dimensions}
(output under \texttt{evaluation\_criteria}):

\begin{itemize}
\item Dimension list (order must not be changed):  
Problem Identification, Problem Restatement, Assumption Formulation,
Model Construction, Model Solving, Code Implementation, Result Analysis.

\item Each dimension must contain \textbf{3--5 specific sub\_criteria}.
Each \texttt{sub\_criteria} must include the following fields:
    \begin{itemize}
      \item \texttt{sub\_criteria}: the evaluation focus (must include task-specific keywords)
      \item \texttt{description}: a description tightly grounded in the semantics of this subtask
      \item \texttt{score}: the integer score assigned to this criterion  
      (the total score within each dimension must equal 100)
      \item \texttt{evaluation\_focus}: a short capability label (no more than six words)
      \item \texttt{scoring\_hint}: concrete, checkable indicators distinguishing high / medium / low scores
    \end{itemize}

\item Rules for writing evaluation criteria:
    \begin{enumerate}
      \item Task-specific keywords must be explicitly reused from \texttt{task\_understanding}.
      \item Generic or vague criteria (e.g., ``whether the goal is clear'') are not allowed.
      \item \texttt{scoring\_hint} should provide operational and verifiable standards;
      quantitative thresholds are preferred whenever possible.
      \item If a dimension is not applicable,
      return an empty list for that dimension
      and add a \texttt{not\_applicable\_reason} explaining why.
    \end{enumerate}

\end{itemize}

\vspace{3mm}
\textbf{III. Output Format and Additional Requirements}:

\begin{itemize}
  \item Only return a single valid JSON object; no extra text is allowed.
  \item The JSON encoding must be UTF-8.
  \item Field order must be preserved
  (\texttt{task\_understanding} first, followed by \texttt{evaluation\_criteria}).
\end{itemize}

\vspace{3mm}
An example output format is shown below:

\begin{verbatim}
```json
{
  "task_understanding": {
    "core_goal": "...",
    "expected_output": "...",
    "key_inputs_constraints": "...",
    "modeling_type": "...",
    "role_in_pipeline": "..."
  },
  "evaluation_criteria": {
    "Problem Identification": [],
    "Problem Restatement": [],
    "Assumption Formulation": [],
    "Model Construction": [],
    "Model Solving": [],
    "Code Implementation": [],
    "Result Analysis": []
  }
}
```
\end{verbatim}

\end{tcolorbox}

%% file: appendix/prompts/eval_prompt.tex
\section{Evaluation Prompt}
\begin{tcolorbox}[
colback=blue!5!white,
colframe=blue!75!black,
title=Stage-wise Evaluation,
fonttitle=\bfseries,
boxrule=0.5mm,
arc=2mm,
left=2mm,
right=2mm,
top=2mm,
bottom=2mm,
breakable
]

\textbf{当前子问题：}

\{\{subproblem\}\}

\vspace{3mm}

\textbf{数学建模论文内容：}

\{\{report\_content\}\}

\vspace{3mm}

\textbf{评分准则（JSON）：}

\begin{verbatim}
```json
{{report_criteria}}
```

\end{verbatim}

\vspace{3mm}
—— 评分任务说明（请严格遵守） ——

\vspace{2mm}

你将作为一名
\textbf{严格、审慎、完全基于证据的数学建模竞赛评审专家}，
依据给定的评分准则，
对上述论文内容进行
\textbf{逐维度、逐评分项的客观评分}。

\vspace{3mm}

\textbf{核心原则（必须遵守）：}

\begin{itemize}
\item \textbf{只依据论文文本，绝不脑补。}
\item 论文未明确写出的内容一律视为不存在。
\item 论文仅模糊提及、缺乏公式、推导、结构说明或方法细节的内容，
视为“部分满足”或“不满足”。
\item 若模型结构、算法流程、关键假设或变量定义未明确给出，
视为“不满足”。
\end{itemize}

\vspace{3mm}

\textbf{逐项评分规则：}

\begin{enumerate}
\item 按照评分准则中给定的顺序，
对每一个评分项分别给出：
\begin{itemize}
\item 评价等级（六级）
\item 得分（必须落在该项规定的分值区间内）
\item 基于论文文本的证据引用
\item 明确判断理由（满足 / 基本满足 / 部分满足 / 基本不满足 / 不满足 / 完全不满足）
\end{itemize}

\item 每一评分项必须明确回答：
\begin{itemize}
\item 论文中是否出现与该评分项直接相关的内容？
\item 是否给出了必要的公式、模型结构、参数含义、推导、算法流程或变量定义？
\item 内容是否完整、严谨、可复现？
\item 若未出现，必须明确说明“论文未出现该内容”，并按规则扣分。
\end{itemize}
\end{enumerate}

\vspace{3mm}

\textbf{六级评价体系（必须严格使用）：}

\begin{enumerate}
\item \textbf{满足（Full）：}
内容完整规范，包含公式、模型、推导与解释，可完整复现。
\item \textbf{基本满足（Almost）：}
核心内容齐全，但细节略有不足，仍可基本复现。
\item \textbf{部分满足（Partial）：}
有相关描述但明显不完整，无法完整复现。
\item \textbf{基本不满足（Barely Not Met）：}
仅有极为浅显的提及，缺乏关键结构与方法细节。
\item \textbf{不满足（Not Met）：}
基本未描述相关内容或明显偏离评分要求。
\item \textbf{完全不满足（Completely Not Met）：}
论文中完全未出现任何相关内容。
\end{enumerate}

\vspace{3mm}

\textbf{判分强制要求：}

\begin{itemize}
\item 得分必须严格落在对应评分项的分值区间内。
\item 评分等级必须与得分保持一致。
\item 不得使用“推测”“可能”“应当”等非证据性表述。
\item 所有评语必须引用论文中出现的真实描述（可节选关键词或短句）。
\end{itemize}

\vspace{3mm}

\textbf{输出格式（仅限 JSON）：}

\begin{verbatim}

```json
{
  "问题识别": [
    {
      "dimension": "[评分维度名称]",
      "comment": "[基于论文文本的评语说明]",
      "score": [得分]
    }
  ],
  "问题复述": [],
  "假设建立": [],
  "模型构建": [],
  "模型求解": [],
  "代码实现": [],
  "结果分析": []
}
```

\end{verbatim}

\end{tcolorbox}

\begin{tcolorbox}[
colback=green!5!white,
colframe=green!75!black,
title=Stage-wise Evaluation (English),
fonttitle=\bfseries,
boxrule=0.5mm,
arc=2mm,
left=2mm,
right=2mm,
top=2mm,
bottom=2mm,
breakable
]

\textbf{Current Subtask:}

\{\{subproblem\}\}

\vspace{3mm}

\textbf{Mathematical Modeling Report Content:}

\{\{report\_content\}\}

\vspace{3mm}

\textbf{Evaluation Criteria (JSON):}

\begin{verbatim}
```json
{{report_criteria}}
```

\end{verbatim}

\vspace{3mm}
—— Evaluation Instructions (Must Be Strictly Followed) ——

\vspace{2mm}

You will act as a
\textbf{strict, cautious, and evidence-based judge in a mathematical modeling competition}.
Based solely on the provided evaluation criteria,
you are required to conduct
\textbf{stage-wise and criterion-wise objective scoring}
of the report above.

\vspace{3mm}

\textbf{Core Principles (Mandatory):}

\begin{itemize}
\item \textbf{Rely exclusively on the report text; do not infer or hallucinate.}
\item Any content not explicitly stated in the report must be treated as nonexistent.
\item Vague mentions without formulas, derivations, structural descriptions, or methodological details
must be considered \emph{partially satisfied} or \emph{not satisfied}.
\item If model structures, algorithmic workflows, key assumptions, or variable definitions
are not explicitly specified, the criterion must be marked as \emph{not satisfied}.
\end{itemize}

\vspace{3mm}

\textbf{Criterion-wise Scoring Rules:}

\begin{enumerate}
\item Follow the order of evaluation criteria strictly.
For each criterion, you must provide:
\begin{itemize}
\item an evaluation level (six-level scale),
\item a numerical score (must fall within the specified score range),
\item explicit evidence cited from the report text,
\item a clear justification (Fully Met / Almost Met / Partially Met / Barely Not Met / Not Met / Completely Not Met).
\end{itemize}

\item For each criterion, you must explicitly address:
\begin{itemize}
\item Whether the report contains content directly relevant to this criterion;
\item Whether necessary elements are provided, including formulas, model structures,
parameter definitions, derivations, algorithmic procedures, or variable specifications;
\item Whether the description is sufficiently complete, rigorous, and reproducible;
\item If absent, explicitly state \emph{``The report does not contain this content''}
and deduct points accordingly.
\end{itemize}
\end{enumerate}

\vspace{3mm}

\textbf{Six-Level Evaluation Scale (Must Be Used Strictly):}

\begin{enumerate}
\item \textbf{Fully Met:}
The report provides a complete and规范 description, including formulas,
models, derivations, and explanations; the method is fully reproducible.
\item \textbf{Almost Met:}
Core components are present, but some details are insufficient;
the method is largely reproducible.
\item \textbf{Partially Met:}
Relevant content is mentioned but clearly incomplete;
the method cannot be fully reproduced.
\item \textbf{Barely Not Met:}
Only superficial mentions are present, with major missing components
such as model structure or algorithmic details.
\item \textbf{Not Met:}
Relevant content is largely absent or deviates substantially
from the criterion requirements.
\item \textbf{Completely Not Met:}
The report contains no information related to this criterion.
\end{enumerate}

\vspace{3mm}

\textbf{Mandatory Scoring Constraints:}

\begin{itemize}
\item Scores must strictly fall within the predefined range of each criterion.
\item Evaluation levels must be consistent with the assigned scores.
\item Do not use speculative language such as `might'', `could'', or ``should''.
\item All comments must cite explicit descriptions from the report text
(keywords or short excerpts are acceptable).
\end{itemize}

\vspace{3mm}

\textbf{Output Format (JSON Only):}

\begin{verbatim}

```json
{
  "Problem Identification": [
    {
      "dimension": "[Criterion name]",
      "comment": "[Evidence-based justification from the report]",
      "score": [Score]
    }
  ],
  "Problem Formulation": [],
  "Assumption Development": [],
  "Model Construction": [],
  "Model Solving": [],
  "Code Implementation": [],
  "Result Analysis": []
}
```

\end{verbatim}

\end{tcolorbox}

%% file: custom.bib
@article{DBLP:journals/corr/abs-2503-05244,
  author       = {Yuning Wu and
                  Jiahao Mei and
                  Ming Yan and
                  Chenliang Li and
                  Shaopeng Lai and
                  Yuran Ren and
                  Zijia Wang and
                  Ji Zhang and
                  Mengyue Wu and
                  Qin Jin and
                  Fei Huang},
  title        = {WritingBench: {A} Comprehensive Benchmark for Generative Writing},
  journal      = {CoRR},
  volume       = {abs/2503.05244},
  year         = {2025},
  url          = {https://doi.org/10.48550/arXiv.2503.05244},
  doi          = {10.48550/ARXIV.2503.05244},
  eprinttype    = {arXiv},
  eprint       = {2503.05244},
  bibsource    = {dblp computer science bibliography, https://dblp.org}
}

@book{banerjee2021mathematical,
  title={Mathematical modeling: models, analysis and applications},
  author={Banerjee, Sandip},
  year={2021},
  publisher={Chapman and Hall/CRC}
}

@misc{deepseekai2024deepseekv32,
      title={DeepSeek-V3.2-Exp: Boosting Long-Context Efficiency with DeepSeek Sparse Attention}, 
      author={DeepSeek-AI},
      year={2025},
}

@article{DBLP:journals/corr/abs-2505-09388,
  author       = {An Yang and
                  Anfeng Li and
                  Baosong Yang and
                  Beichen Zhang and
                  Binyuan Hui and
                  Bo Zheng and
                  Bowen Yu and
                  Chang Gao and
                  Chengen Huang and
                  Chenxu Lv and
                  Chujie Zheng and
                  Dayiheng Liu and
                  Fan Zhou and
                  Fei Huang and
                  Feng Hu and
                  Hao Ge and
                  Haoran Wei and
                  Huan Lin and
                  Jialong Tang and
                  Jian Yang and
                  Jianhong Tu and
                  Jianwei Zhang and
                  Jian Yang and
                  Jiaxi Yang and
                  Jingren Zhou and
                  Junyang Lin and
                  Kai Dang and
                  Keqin Bao and
                  Kexin Yang and
                  Le Yu and
                  Lianghao Deng and
                  Mei Li and
                  Mingfeng Xue and
                  Mingze Li and
                  Pei Zhang and
                  Peng Wang and
                  Qin Zhu and
                  Rui Men and
                  Ruize Gao and
                  Shixuan Liu and
                  Shuang Luo and
                  Tianhao Li and
                  Tianyi Tang and
                  Wenbiao Yin and
                  Xingzhang Ren and
                  Xinyu Wang and
                  Xinyu Zhang and
                  Xuancheng Ren and
                  Yang Fan and
                  Yang Su and
                  Yichang Zhang and
                  Yinger Zhang and
                  Yu Wan and
                  Yuqiong Liu and
                  Zekun Wang and
                  Zeyu Cui and
                  Zhenru Zhang and
                  Zhipeng Zhou and
                  Zihan Qiu},
  title        = {Qwen3 Technical Report},
  journal      = {CoRR},
  volume       = {abs/2505.09388},
  year         = {2025},
  url          = {https://doi.org/10.48550/arXiv.2505.09388},
  doi          = {10.48550/ARXIV.2505.09388},
  eprinttype    = {arXiv},
  eprint       = {2505.09388},
  bibsource    = {dblp computer science bibliography, https://dblp.org}
}

@article{DBLP:journals/corr/abs-2412-15115,
  author       = {An Yang and
                  Baosong Yang and
                  Beichen Zhang and
                  Binyuan Hui and
                  Bo Zheng and
                  Bowen Yu and
                  Chengyuan Li and
                  Dayiheng Liu and
                  Fei Huang and
                  Haoran Wei and
                  Huan Lin and
                  Jian Yang and
                  Jianhong Tu and
                  Jianwei Zhang and
                  Jianxin Yang and
                  Jiaxi Yang and
                  Jingren Zhou and
                  Junyang Lin and
                  Kai Dang and
                  Keming Lu and
                  Keqin Bao and
                  Kexin Yang and
                  Le Yu and
                  Mei Li and
                  Mingfeng Xue and
                  Pei Zhang and
                  Qin Zhu and
                  Rui Men and
                  Runji Lin and
                  Tianhao Li and
                  Tingyu Xia and
                  Xingzhang Ren and
                  Xuancheng Ren and
                  Yang Fan and
                  Yang Su and
                  Yichang Zhang and
                  Yu Wan and
                  Yuqiong Liu and
                  Zeyu Cui and
                  Zhenru Zhang and
                  Zihan Qiu},
  title        = {Qwen2.5 Technical Report},
  journal      = {CoRR},
  volume       = {abs/2412.15115},
  year         = {2024},
  url          = {https://doi.org/10.48550/arXiv.2412.15115},
  doi          = {10.48550/ARXIV.2412.15115},
  eprinttype    = {arXiv},
  eprint       = {2412.15115},
  bibsource    = {dblp computer science bibliography, https://dblp.org}
}

@article{DBLP:journals/corr/abs-2502-13923,
  author       = {Shuai Bai and
                  Keqin Chen and
                  Xuejing Liu and
                  Jialin Wang and
                  Wenbin Ge and
                  Sibo Song and
                  Kai Dang and
                  Peng Wang and
                  Shijie Wang and
                  Jun Tang and
                  Humen Zhong and
                  Yuanzhi Zhu and
                  Ming{-}Hsuan Yang and
                  Zhaohai Li and
                  Jianqiang Wan and
                  Pengfei Wang and
                  Wei Ding and
                  Zheren Fu and
                  Yiheng Xu and
                  Jiabo Ye and
                  Xi Zhang and
                  Tianbao Xie and
                  Zesen Cheng and
                  Hang Zhang and
                  Zhibo Yang and
                  Haiyang Xu and
                  Junyang Lin},
  title        = {Qwen2.5-VL Technical Report},
  journal      = {CoRR},
  volume       = {abs/2502.13923},
  year         = {2025},
  url          = {https://doi.org/10.48550/arXiv.2502.13923},
  doi          = {10.48550/ARXIV.2502.13923},
  eprinttype    = {arXiv},
  eprint       = {2502.13923},
  bibsource    = {dblp computer science bibliography, https://dblp.org}
}

@article{weir2005quantifying,
  title={Quantifying test-retest reliability using the intraclass correlation coefficient and the SEM},
  author={Weir, Joseph P},
  journal={The Journal of Strength \& Conditioning Research},
  volume={19},
  number={1},
  pages={231--240},
  year={2005},
  publisher={LWW}
}

@article{DBLP:journals/corr/abs-2505-09970,
  author       = {Mrinal Rawat and
                  Ambuje Gupta and
                  Rushil Goomer and
                  Alessandro Di Bari and
                  Neha Gupta and
                  Roberto Pieraccini},
  title        = {Pre-Act: Multi-Step Planning and Reasoning Improves Acting in {LLM}
                  Agents},
  journal      = {CoRR},
  volume       = {abs/2505.09970},
  year         = {2025},
  url          = {https://doi.org/10.48550/arXiv.2505.09970},
  doi          = {10.48550/ARXIV.2505.09970},
  eprinttype    = {arXiv},
  eprint       = {2505.09970},
  bibsource    = {dblp computer science bibliography, https://dblp.org}
}

@article{DBLP:journals/corr/abs-2511-08151,
  author       = {Xuchen Li and
                  Ruitao Wu and
                  Xuanbo Liu and
                  Xukai Wang and
                  Jinbo Hu and
                  Zhixin Bai and
                  Bohan Zeng and
                  Hao Liang and
                  Leheng Chen and
                  Mingrui Chen and
                  Haitian Zhong and
                  Xuanlin Yang and
                  Xu{-}Yao Zhang and
                  Liu Liu and
                  Jia Li and
                  Kaiqi Huang and
                  Jiahao Xu and
                  Haitao Mi and
                  Wentao Zhang and
                  Bin Dong},
  title        = {SciAgent: {A} Unified Multi-Agent System for Generalistic Scientific
                  Reasoning},
  journal      = {CoRR},
  volume       = {abs/2511.08151},
  year         = {2025},
  url          = {https://doi.org/10.48550/arXiv.2511.08151},
  doi          = {10.48550/ARXIV.2511.08151},
  eprinttype    = {arXiv},
  eprint       = {2511.08151},
  bibsource    = {dblp computer science bibliography, https://dblp.org}
}

@article{DBLP:journals/corr/abs-2506-14683,
  author       = {Leonhard Applis and
                  Yuntong Zhang and
                  Shanchao Liang and
                  Nan Jiang and
                  Lin Tan and
                  Abhik Roychoudhury},
  title        = {Unified Software Engineering agent as {AI} Software Engineer},
  journal      = {CoRR},
  volume       = {abs/2506.14683},
  year         = {2025},
  url          = {https://doi.org/10.48550/arXiv.2506.14683},
  doi          = {10.48550/ARXIV.2506.14683},
  eprinttype    = {arXiv},
  eprint       = {2506.14683},
  bibsource    = {dblp computer science bibliography, https://dblp.org}
}

@inproceedings{DBLP:conf/iclr/JimenezYWYPPN24,
  author       = {Carlos E. Jimenez and
                  John Yang and
                  Alexander Wettig and
                  Shunyu Yao and
                  Kexin Pei and
                  Ofir Press and
                  Karthik R. Narasimhan},
  title        = {SWE-bench: Can Language Models Resolve Real-world Github Issues?},
  booktitle    = {The Twelfth International Conference on Learning Representations,
                  {ICLR} 2024, Vienna, Austria, May 7-11, 2024},
  publisher    = {OpenReview.net},
  year         = {2024},
  url          = {https://openreview.net/forum?id=VTF8yNQM66},
  bibsource    = {dblp computer science bibliography, https://dblp.org}
}

@article{DBLP:journals/corr/abs-2505-14615,
  author       = {Anjiang Wei and
                  Yuheng Wu and
                  Yingjia Wan and
                  Tarun Suresh and
                  Huanmi Tan and
                  Zhanke Zhou and
                  Sanmi Koyejo and
                  Ke Wang and
                  Alex Aiken},
  title        = {SATBench: Benchmarking LLMs' Logical Reasoning via Automated
                  Puzzle Generation from {SAT} Formulas},
  journal      = {CoRR},
  volume       = {abs/2505.14615},
  year         = {2025},
  url          = {https://doi.org/10.48550/arXiv.2505.14615},
  doi          = {10.48550/ARXIV.2505.14615},
  eprinttype    = {arXiv},
  eprint       = {2505.14615},
  bibsource    = {dblp computer science bibliography, https://dblp.org}
}

@misc{song2025evaluatinglargelanguagemodels,
      title={Evaluating Large Language Models in Scientific Discovery}, 
      author={Zhangde Song and Jieyu Lu and Yuanqi Du and Botao Yu and Thomas M. Pruyn and Yue Huang and Kehan Guo and Xiuzhe Luo and Yuanhao Qu and Yi Qu and Yinkai Wang and Haorui Wang and Jeff Guo and Jingru Gan and Parshin Shojaee and Di Luo and Andres M Bran and Gen Li and Qiyuan Zhao and Shao-Xiong Lennon Luo and Yuxuan Zhang and Xiang Zou and Wanru Zhao and Yifan F. Zhang and Wucheng Zhang and Shunan Zheng and Saiyang Zhang and Sartaaj Takrim Khan and Mahyar Rajabi-Kochi and Samantha Paradi-Maropakis and Tony Baltoiu and Fengyu Xie and Tianyang Chen and Kexin Huang and Weiliang Luo and Meijing Fang and Xin Yang and Lixue Cheng and Jiajun He and Soha Hassoun and Xiangliang Zhang and Wei Wang and Chandan K. Reddy and Chao Zhang and Zhiling Zheng and Mengdi Wang and Le Cong and Carla P. Gomes and Chang-Yu Hsieh and Aditya Nandy and Philippe Schwaller and Heather J. Kulik and Haojun Jia and Huan Sun and Seyed Mohamad Moosavi and Chenru Duan},
      year={2025},
      eprint={2512.15567},
      archivePrefix={arXiv},
      primaryClass={cs.AI},
      url={https://arxiv.org/abs/2512.15567}, 
}

@misc{AIME,
  author = {AIME},
  title = {Aime problems and solutions},
  url = {https://artofproblemsolving.com/wiki/index.php/AIME_Problems_and_Solutions},
  year = {2025}
}

@article{shrout1979intraclass,
  title={Intraclass correlations: uses in assessing rater reliability.},
  author={Shrout, Patrick E and Fleiss, Joseph L},
  journal={Psychological bulletin},
  volume={86},
  number={2},
  pages={420},
  year={1979},
  publisher={American Psychological Association}
}

@online{GMCM2025Awards,
  title        = {“华为杯”第二十二届中国研究生数学建模竞赛获奖名单公布
                  [Award List of the 22nd China Graduate Mathematical Contest in Modeling]},
  author       = {{China Graduate Mathematical Contest in Modeling Organizing Committee}},
  date         = {2025-11-24},
  url          = {https://cpipc.acge.org.cn/cw/hp/4},
  urldate      = {2025-12-01},
  note         = {Official announcement on the China Graduate Innovation Practice Competition website (in Chinese).}
}

@book{bender2000introduction,
  title={An introduction to mathematical modeling},
  author={Bender, Edward A},
  year={2000},
  publisher={Courier Corporation}
}

@misc{o4mini,
  author       = {{OpenAI}},
  title        = {Introducing o3 and o4-mini},
  howpublished = {\url{https://openai.com/index/introducing-o3-and-o4-mini/}},
  year         = {2025}
}

@article{xi2025information,
  title={Information Bottleneck-Enhanced Reinforcement Learning for Solving Operation Research Problems},
  author={Xi, Ruozhang and Ni, Yao and Wu, Wangyu},
  journal={Sensors},
  volume={25},
  number={24},
  pages={7572},
  year={2025},
  publisher={MDPI}
}

@article{DBLP:journals/corr/abs-2509-17677,
  author       = {Xiyuan Zhou and
                  Xinlei Wang and
                  Yirui He and
                  Yang Wu and
                  Ruixi Zou and
                  Yuheng Cheng and
                  Yulu Xie and
                  Wenxuan Liu and
                  Huan Zhao and
                  Yan Xu and
                  Jinjin Gu and
                  Junhua Zhao},
  title        = {EngiBench: {A} Benchmark for Evaluating Large Language Models on Engineering
                  Problem Solving},
  journal      = {CoRR},
  volume       = {abs/2509.17677},
  year         = {2025},
  url          = {https://doi.org/10.48550/arXiv.2509.17677},
  doi          = {10.48550/ARXIV.2509.17677},
  eprinttype    = {arXiv},
  eprint       = {2509.17677},
  bibsource    = {dblp computer science bibliography, https://dblp.org}
}

@article{DBLP:journals/corr/abs-2506-16395,
  author       = {Zhexu Wang and
                  Yiping Liu and
                  Yejie Wang and
                  Wenyang He and
                  Bofei Gao and
                  Muxi Diao and
                  Yanxu Chen and
                  Kelin Fu and
                  Flood Sung and
                  Zhilin Yang and
                  Tianyu Liu and
                  Weiran Xu},
  title        = {OJBench: {A} Competition Level Code Benchmark For Large Language Models},
  journal      = {CoRR},
  volume       = {abs/2506.16395},
  year         = {2025},
  url          = {https://doi.org/10.48550/arXiv.2506.16395},
  doi          = {10.48550/ARXIV.2506.16395},
  eprinttype    = {arXiv},
  eprint       = {2506.16395},
  bibsource    = {dblp computer science bibliography, https://dblp.org}
}

@article{DBLP:journals/corr/abs-2508-09101,
  author       = {Jason Chou and
                  Ao Liu and
                  Yuchi Deng and
                  Zhiying Zeng and
                  Tao Zhang and
                  Haotian Zhu and
                  Jianwei Cai and
                  Yue Mao and
                  Chenchen Zhang and
                  Lingyun Tan and
                  Ziyan Xu and
                  Bohui Zhai and
                  Hengyi Liu and
                  Speed Zhu and
                  Wiggin Zhou and
                  Fengzong Lian},
  title        = {AutoCodeBench: Large Language Models are Automatic Code Benchmark
                  Generators},
  journal      = {CoRR},
  volume       = {abs/2508.09101},
  year         = {2025},
  url          = {https://doi.org/10.48550/arXiv.2508.09101},
  doi          = {10.48550/ARXIV.2508.09101},
  eprinttype    = {arXiv},
  eprint       = {2508.09101},
  bibsource    = {dblp computer science bibliography, https://dblp.org}
}

@article{DBLP:journals/corr/abs-2303-08774,
  author       = {OpenAI},
  title        = {{GPT-4} Technical Report},
  journal      = {CoRR},
  volume       = {abs/2303.08774},
  year         = {2023},
  url          = {https://doi.org/10.48550/arXiv.2303.08774},
  doi          = {10.48550/ARXIV.2303.08774},
  eprinttype    = {arXiv},
  eprint       = {2303.08774},
  bibsource    = {dblp computer science bibliography, https://dblp.org}
}

@techreport{grok,
  title       = {Grok 4 Model Card},
  author      = {xAI},
  year        = {2025},
  institution = {xAI},
  url         = {https://data.x.ai/2025-08-20-grok-4-model-card.pdf}
}

@techreport{OpenAI2025GPT5SystemCard,
 title       = {GPT-5 System Card},
  author      = {OpenAI},
  year        = {2025},
  institution = {OpenAI},
  url         = {https://openai.com/index/gpt-5-system-card/}
}

@article{DBLP:journals/corr/abs-2507-06261,
  author       = {Gemini Team},
  title        = {Gemini 2.5: Pushing the Frontier with Advanced Reasoning, Multimodality,
                  Long Context, and Next Generation Agentic Capabilities},
  journal      = {CoRR},
  volume       = {abs/2507.06261},
  year         = {2025},
  url          = {https://doi.org/10.48550/arXiv.2507.06261},
  doi          = {10.48550/ARXIV.2507.06261},
  eprinttype    = {arXiv},
  eprint       = {2507.06261},
  bibsource    = {dblp computer science bibliography, https://dblp.org}
}

@misc{deepseekai2025deepseekv32pushingfrontieropen,
      title={DeepSeek-V3.2: Pushing the Frontier of Open Large Language Models}, 
      author={DeepSeek-AI and Aixin Liu and Aoxue Mei and Bangcai Lin and Bing Xue and Bingxuan Wang and Bingzheng Xu and Bochao Wu and Bowei Zhang and Chaofan Lin and Chen Dong and Chengda Lu and Chenggang Zhao and Chengqi Deng and Chenhao Xu and Chong Ruan and Damai Dai and Daya Guo and Dejian Yang and Deli Chen and Erhang Li and Fangqi Zhou and Fangyun Lin and Fucong Dai and Guangbo Hao and Guanting Chen and Guowei Li and H. Zhang and Hanwei Xu and Hao Li and Haofen Liang and Haoran Wei and Haowei Zhang and Haowen Luo and Haozhe Ji and Honghui Ding and Hongxuan Tang and Huanqi Cao and Huazuo Gao and Hui Qu and Hui Zeng and Jialiang Huang and Jiashi Li and Jiaxin Xu and Jiewen Hu and Jingchang Chen and Jingting Xiang and Jingyang Yuan and Jingyuan Cheng and Jinhua Zhu and Jun Ran and Junguang Jiang and Junjie Qiu and Junlong Li and Junxiao Song and Kai Dong and Kaige Gao and Kang Guan and Kexin Huang and Kexing Zhou and Kezhao Huang and Kuai Yu and Lean Wang and Lecong Zhang and Lei Wang and Liang Zhao and Liangsheng Yin and Lihua Guo and Lingxiao Luo and Linwang Ma and Litong Wang and Liyue Zhang and M. S. Di and M. Y Xu and Mingchuan Zhang and Minghua Zhang and Minghui Tang and Mingxu Zhou and Panpan Huang and Peixin Cong and Peiyi Wang and Qiancheng Wang and Qihao Zhu and Qingyang Li and Qinyu Chen and Qiushi Du and Ruiling Xu and Ruiqi Ge and Ruisong Zhang and Ruizhe Pan and Runji Wang and Runqiu Yin and Runxin Xu and Ruomeng Shen and Ruoyu Zhang and S. H. Liu and Shanghao Lu and Shangyan Zhou and Shanhuang Chen and Shaofei Cai and Shaoyuan Chen and Shengding Hu and Shengyu Liu and Shiqiang Hu and Shirong Ma and Shiyu Wang and Shuiping Yu and Shunfeng Zhou and Shuting Pan and Songyang Zhou and Tao Ni and Tao Yun and Tian Pei and Tian Ye and Tianyuan Yue and Wangding Zeng and Wen Liu and Wenfeng Liang and Wenjie Pang and Wenjing Luo and Wenjun Gao and Wentao Zhang and Xi Gao and Xiangwen Wang and Xiao Bi and Xiaodong Liu and Xiaohan Wang and Xiaokang Chen and Xiaokang Zhang and Xiaotao Nie and Xin Cheng and Xin Liu and Xin Xie and Xingchao Liu and Xingkai Yu and Xingyou Li and Xinyu Yang and Xinyuan Li and Xu Chen and Xuecheng Su and Xuehai Pan and Xuheng Lin and Xuwei Fu and Y. Q. Wang and Yang Zhang and Yanhong Xu and Yanru Ma and Yao Li and Yao Li and Yao Zhao and Yaofeng Sun and Yaohui Wang and Yi Qian and Yi Yu and Yichao Zhang and Yifan Ding and Yifan Shi and Yiliang Xiong and Ying He and Ying Zhou and Yinmin Zhong and Yishi Piao and Yisong Wang and Yixiao Chen and Yixuan Tan and Yixuan Wei and Yiyang Ma and Yiyuan Liu and Yonglun Yang and Yongqiang Guo and Yongtong Wu and Yu Wu and Yuan Cheng and Yuan Ou and Yuanfan Xu and Yuduan Wang and Yue Gong and Yuhan Wu and Yuheng Zou and Yukun Li and Yunfan Xiong and Yuxiang Luo and Yuxiang You and Yuxuan Liu and Yuyang Zhou and Z. F. Wu and Z. Z. Ren and Zehua Zhao and Zehui Ren and Zhangli Sha and Zhe Fu and Zhean Xu and Zhenda Xie and Zhengyan Zhang and Zhewen Hao and Zhibin Gou and Zhicheng Ma and Zhigang Yan and Zhihong Shao and Zhixian Huang and Zhiyu Wu and Zhuoshu Li and Zhuping Zhang and Zian Xu and Zihao Wang and Zihui Gu and Zijia Zhu and Zilin Li and Zipeng Zhang and Ziwei Xie and Ziyi Gao and Zizheng Pan and Zongqing Yao and Bei Feng and Hui Li and J. L. Cai and Jiaqi Ni and Lei Xu and Meng Li and Ning Tian and R. J. Chen and R. L. Jin and S. S. Li and Shuang Zhou and Tianyu Sun and X. Q. Li and Xiangyue Jin and Xiaojin Shen and Xiaosha Chen and Xinnan Song and Xinyi Zhou and Y. X. Zhu and Yanping Huang and Yaohui Li and Yi Zheng and Yuchen Zhu and Yunxian Ma and Zhen Huang and Zhipeng Xu and Zhongyu Zhang and Dongjie Ji and Jian Liang and Jianzhong Guo and Jin Chen and Leyi Xia and Miaojun Wang and Mingming Li and Peng Zhang and Ruyi Chen and Shangmian Sun and Shaoqing Wu and Shengfeng Ye and T. Wang and W. L. Xiao and Wei An and Xianzu Wang and Xiaowen Sun and Xiaoxiang Wang and Ying Tang and Yukun Zha and Zekai Zhang and Zhe Ju and Zhen Zhang and Zihua Qu},
      year={2025},
      eprint={2512.02556},
      archivePrefix={arXiv},
      primaryClass={cs.CL},
      url={https://arxiv.org/abs/2512.02556}, 
}

@inproceedings{DBLP:conf/nips/HendrycksBKABTS21,
  author       = {Dan Hendrycks and
                  Collin Burns and
                  Saurav Kadavath and
                  Akul Arora and
                  Steven Basart and
                  Eric Tang and
                  Dawn Song and
                  Jacob Steinhardt},
  editor       = {Joaquin Vanschoren and
                  Sai{-}Kit Yeung},
  title        = {Measuring Mathematical Problem Solving With the {MATH} Dataset},
  booktitle    = {Proceedings of the Neural Information Processing Systems Track on
                  Datasets and Benchmarks 1, NeurIPS Datasets and Benchmarks 2021, December
                  2021, virtual},
  year         = {2021},
  url          = {https://datasets-benchmarks-proceedings.neurips.cc/paper/2021/hash/be83ab3ecd0db773eb2dc1b0a17836a1-Abstract-round2.html},
  bibsource    = {dblp computer science bibliography, https://dblp.org}
}

@inproceedings{DBLP:conf/iclr/Yang0HGSHFSL025,
  author       = {Zhicheng Yang and
                  Yiwei Wang and
                  Yinya Huang and
                  Zhijiang Guo and
                  Wei Shi and
                  Xiongwei Han and
                  Liang Feng and
                  Linqi Song and
                  Xiaodan Liang and
                  Jing Tang},
  title        = {OptiBench Meets ReSocratic: Measure and Improve LLMs for Optimization
                  Modeling},
  booktitle    = {The Thirteenth International Conference on Learning Representations,
                  {ICLR} 2025, Singapore, April 24-28, 2025},
  publisher    = {OpenReview.net},
  year         = {2025},
  url          = {https://openreview.net/forum?id=fsDZwS49uY},
  bibsource    = {dblp computer science bibliography, https://dblp.org}
}

@article{DBLP:journals/corr/abs-2505-14148,
  author       = {Fan Liu and
                  Zherui Yang and
                  Cancheng Liu and
                  Tianrui Song and
                  Xiaofeng Gao and
                  Hao Liu},
  title        = {MM-Agent: {LLM} as Agents for Real-world Mathematical Modeling Problem},
  journal      = {CoRR},
  volume       = {abs/2505.14148},
  year         = {2025},
  url          = {https://doi.org/10.48550/arXiv.2505.14148},
  doi          = {10.48550/ARXIV.2505.14148},
  eprinttype    = {arXiv},
  eprint       = {2505.14148},
  bibsource    = {dblp computer science bibliography, https://dblp.org}
}

@inproceedings{DBLP:conf/acl/YanFYX00Z25,
  author       = {Xiangchao Yan and
                  Shiyang Feng and
                  Jiakang Yuan and
                  Renqiu Xia and
                  Bin Wang and
                  Lei Bai and
                  Bo Zhang},
  editor       = {Wanxiang Che and
                  Joyce Nabende and
                  Ekaterina Shutova and
                  Mohammad Taher Pilehvar},
  title        = {{SURVEYFORGE} : On the Outline Heuristics, Memory-Driven Generation,
                  and Multi-dimensional Evaluation for Automated Survey Writing},
  booktitle    = {Proceedings of the 63rd Annual Meeting of the Association for Computational
                  Linguistics (Volume 1: Long Papers), {ACL} 2025, Vienna, Austria,
                  July 27 - August 1, 2025},
  pages        = {12444--12465},
  publisher    = {Association for Computational Linguistics},
  year         = {2025},
  url          = {https://aclanthology.org/2025.acl-long.609/},
  bibsource    = {dblp computer science bibliography, https://dblp.org}
}

@misc{wang2025ariseagenticrubricguidediterative,
      title={ARISE: Agentic Rubric-Guided Iterative Survey Engine for Automated Scholarly Paper Generation}, 
      author={Zi Wang and Xingqiao Wang and Sangah Lee and Xiaowei Xu},
      year={2025},
      eprint={2511.17689},
      archivePrefix={arXiv},
      primaryClass={cs.DL},
      url={https://arxiv.org/abs/2511.17689}, 
}

@inproceedings{DBLP:conf/kdd/MohammadiLLY25,
  author       = {Mahmoud Mohammadi and
                  Yipeng Li and
                  Jane Lo and
                  Wendy Yip},
  editor       = {Luiza Antonie and
                  Jian Pei and
                  Xiaohui Yu and
                  Flavio Chierichetti and
                  Hady W. Lauw and
                  Yizhou Sun and
                  Srinivasan Parthasarathy},
  title        = {Evaluation and Benchmarking of {LLM} Agents: {A} Survey},
  booktitle    = {Proceedings of the 31st {ACM} {SIGKDD} Conference on Knowledge Discovery
                  and Data Mining, V.2, {KDD} 2025, Toronto ON, Canada, August 3-7,
                  2025},
  pages        = {6129--6139},
  publisher    = {{ACM}},
  year         = {2025},
  url          = {https://doi.org/10.1145/3711896.3736570},
  doi          = {10.1145/3711896.3736570},
  bibsource    = {dblp computer science bibliography, https://dblp.org}
}

@misc{qian2025modelingagentbridgingllmsmathematical,
      title={ModelingAgent: Bridging LLMs and Mathematical Modeling for Real-World Challenges}, 
      author={Cheng Qian and Hongyi Du and Hongru Wang and Xiusi Chen and Yuji Zhang and Avirup Sil and Chengxiang Zhai and Kathleen McKeown and Heng Ji},
      year={2025},
      eprint={2505.15068},
      archivePrefix={arXiv},
      primaryClass={cs.AI},
      url={https://arxiv.org/abs/2505.15068}, 
}
